\pdfoutput=1
\documentclass[11pt,dvipsnames]{article}

\usepackage[final]{acl}

\usepackage{times}
\usepackage{enumitem}
\usepackage{lipsum}
\usepackage{latexsym}

\usepackage[T1]{fontenc}
\usepackage[utf8]{inputenc}

\usepackage{microtype}

\usepackage{inconsolata}

\usepackage{graphicx}
\usepackage{multirow}
\usepackage{adjustbox}

\usepackage{amsmath}
\usepackage{booktabs}
\usepackage{tabularx}
\usepackage{enumitem}
\usepackage[table]{colortbl}
\usepackage{dirtree}
\usepackage{tabto}
\usepackage{hyperref}

\title{Tree-of-Debate: Multi-Persona Debate Trees Elicit Critical Thinking for Scientific Comparative Analysis}

  \author{Priyanka Kargupta, Ishika Agarwal, Tal August, Jiawei Han\\
  Department of Computer Science, University of Illinois at Urbana-Champaign\\
  \texttt{\{pk36, ishikaa2, taugust, hanj\}@illinois.edu} \\
}

\begin{document}
\maketitle

\begin{abstract}
With the exponential growth of research facilitated by modern technology and improved accessibility, scientific discoveries have become increasingly fragmented within and across fields. This makes it challenging to assess the significance, novelty, incremental findings, and equivalent ideas between related works, particularly those from different research communities. Large language models (LLMs) have recently demonstrated strong quantitative and qualitative reasoning abilities, and multi-agent LLM debates have shown promise in handling complex reasoning tasks by exploring diverse perspectives and reasoning paths. Inspired by this, we introduce \textbf{Tree-of-Debate} (ToD), a framework which converts scientific papers into LLM personas that debate their respective novelties. To emphasize structured, critical reasoning rather than focusing solely on outcomes, ToD dynamically constructs a debate tree, enabling fine-grained analysis of independent novelty arguments within scholarly articles. Through experiments on scientific literature across various domains, evaluated by expert researchers, we demonstrate that ToD generates informative arguments, effectively contrasts papers, and supports researchers in their literature review.
\end{abstract}

\section{Introduction}
\begin{figure}
        \centering
        \includegraphics[width=1.0\linewidth]{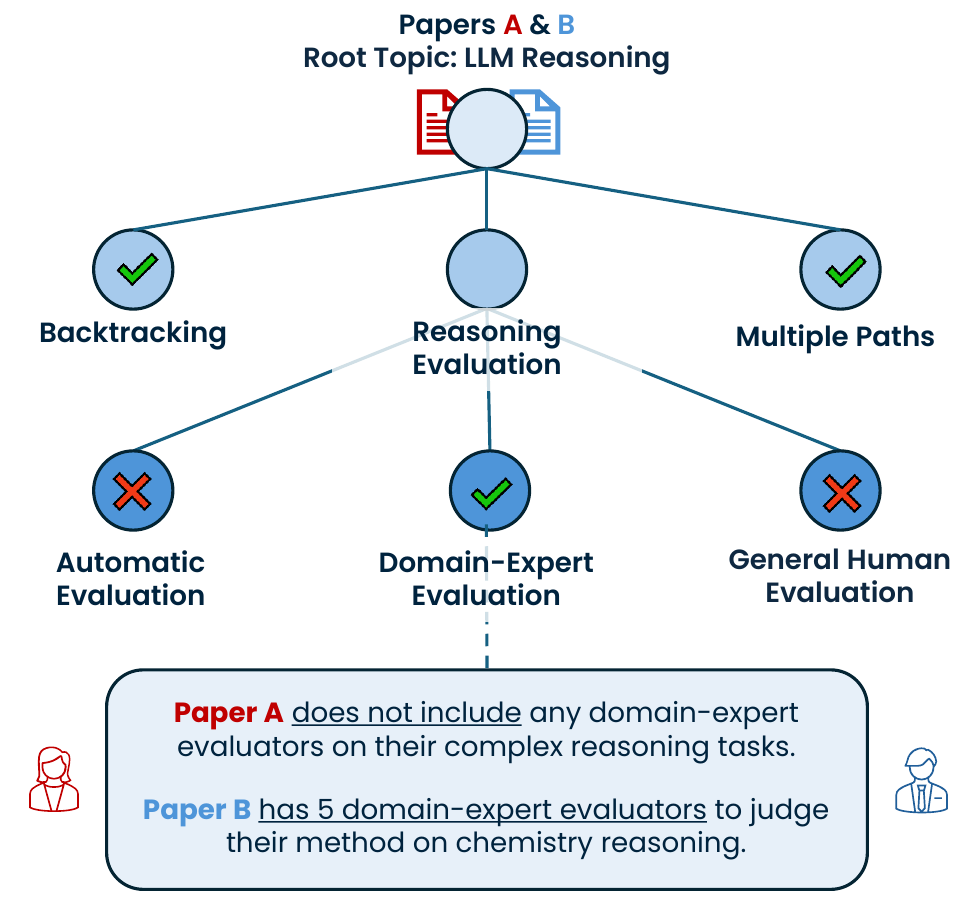}
        \caption{A hierarchy of contributions made by Papers $A$ and $B$, specific to the root topic. Green check mark: a single paper makes a unique contribution; red X: an overlapping contribution.}
        \label{fig:hierarchy}
    \end{figure}
    
\par Navigating and identifying new and relevant research findings has become non-trivial with the popularity of open-access repositories. For example, arXiv received over 24,000 submissions in October 2024 \cite{arxiv_monthly_submissions}, inundating researchers with an overwhelming volume of information. This astronomical surge in scholarly articles makes it difficult to identify novel findings and discern the distinctions between related papers, especially those presenting similar ideas from different angles (e.g., papers from different research communities).

\par Automatically generating comparative summaries of research papers has proven valuable for addressing these challenges \cite{whats_new}. Existing comparative summarization works \cite{Strhle2023} typically follow a two-step pipeline: (1) construct extractive summaries for each document to (2) identify their similarities and differences \cite{lerman2009contrastive,gunel2024strum}. However, despite using large language models (LLMs), these methods often focus on surface-level semantic differences, which may not capture the most relevant distinctions. For example, when comparing pre-trained models like ``BERT'' \cite{devlin-etal-2019-bert} and ``RoBERTa'' \cite{Liu2019RoBERTaAR}, it is crucial to note that RoBERTa omits next-sentence prediction, trains on ten times more data, and achieves superior performance. These insights require complex, comparative reasoning beyond basic semantics, as they rely on understanding BERT's contributions \textit{in the context} of RoBERTa's. Thus, we propose the following principles:

% need for debate
\par{\textbf{Multi-persona debates elicit complex, comparative reasoning.}} We explore the use of multi-agent debates for inducing fine-grained, comparative reasoning. These debates simulate group discussions where agents suggest diverse answers, critique one another, and refine responses to produce better outputs \cite{chan2023chateval, liang2023encouraging}. Recent work has also introduced defining LLM agents as personas with distinct characteristics or values, enabling them to generate outputs that reflect the diverse perspectives needed to solve multi-faceted problems \cite{tseng2024two,wang-etal-2024-unleashing}. Inspired by this, we propose \textit{converting scientific \underline{papers into personas} that debate} each other to foster critical analysis. For instance, while the papers debate their respective contributions to a topic, they critically evaluate each other's novelty and significance \textit{relative to their own claims}.

% need for structured debate
\par{\textbf{Tree-structured debates allow for independent assessments of different contributions at varying depths.}} A scientific paper often makes contributions (e.g., methodology, dataset, evaluation metric) that can be deconstructed into multiple ``sub-ideas.'' Some sub-ideas may or may not be novel (e.g., uses an existing architecture, but proposes novel fine-tuning and evaluation mechanisms) and consequently, should be independently evaluated for their \textit{degree} of novelty. Hence, an unstructured debate combining all ideas is insufficient for handling the complexity of scientific comparative analysis. We instead propose a \textit{tree-structured} debate, where each \textit{node} represents a specific contribution topic being debated, and an \textit{edge} indicates unresolved points or interesting questions from the parent debate node which warrant further exploration in a child node. Figure \ref{fig:hierarchy} illustrates these topical relationships.

\par{\textbf{Iterative retrieval throughout a debate improves fine-grained reasoning.}} Due to their lengthy nature, providing an entire paper in-context is ineffective, as details specific to the debate node topic may be overshadowed \cite{li-etal-2024-loogle}. Conversely, using only the title and abstract results in high-level comparisons based on surface-level semantic differences. To address these long-context challenges, we propose an \textit{iterative retrieval} process, where retrieval queries are dynamically determined by the debate's content. This ensures the retrieved content is targeted to the specific contribution in question, enabling personas to generate more compelling affirmative or opposing arguments. For instance, as the debate progresses from ``reasoning evaluation'' to ``domain-expert evaluation'' in Figure \ref{fig:hierarchy}, the evidence pool is updated to be more fine-grained and relevant to the subtopic.

% multiple contributions -> need for STRUCTURED debate

\par We integrate these proposed principles into \textbf{Tree-of-Debate}, a framework which dynamically structures a debate between paper personas, conducted by a moderator. First, each persona prepares (\textbf{self-deliberation}) by retrieving topic-relevant segments from their paper, identifying their novel contributions, and updating their evidence pool based on the opposition's claimed contributions (Section \ref{sec:self-deliberate}). Based on this, the moderator determines the most valuable subtopics to explore (e.g., second level of Figure \ref{fig:hierarchy}). For each subtopic, a \textbf{child debate node is formed}, where each persona presents their initial arguments, responds to one another (e.g., clarifying questions, doubts), and revises their argument based on the interaction. Based on the debate, the moderator determines if the \textbf{debate node is worth expanding} and exploring deeper into (Section \ref{sec:expansion}). If so, a more fine-grained set of subtopics are determined for the next level of debate children (Section \ref{sec:debate_round_format}). Our contributions can be summarized as: 
\begin{itemize}
    \itemsep-0.1em
    \item We introduce \textbf{Tree-of-Debate}, a structured multi-persona debate framework, to generate fine-grained contrastive summaries.  
    \item Tree-of-Debate can \textbf{dynamically construct a debate tree} to reason about fine-grained arguments discussed in scholarly articles. 
    \item Through experiments on \textbf{real-world scientific literature}, we show that Tree-of-Debates elicits informative arguments and contrasts papers, aiding researchers in their work. 
\end{itemize}

\textbf{Reproducibility:} We provide our dataset and source code\footnote{\url{https://github.com/pkargupta/tree-of-debate}} to facilitate further studies.

\section{Related Work}
\subsection{Persona Creation \& Debate} Similar to how a person's background shapes their abilities, recent work has explored assigning personas to LLMs to capture diverse perspectives and extract unique capabilities \cite{self-play}. For instance, \citet{bridger} creates author personas for author recommendation by applying named entity recognition to papers and matches authors based on shared terminology. As we aim to highlight specific novelties and incremental contributions between two papers, we instead define a \textit{paper persona}. While \citet{bridger}'s personas represent the author's multiple works and are used solely for comparison, ours represent the paper, actively debating for and defending it. Other studies have also leveraged persona-driven debate by assigning multiple personas-- such as affirmative and negative debaters along with a judge--- to synthesize diverse reasoning steps for tasks like commonsense generation and arithmetic, thereby reducing confirmation bias inherent in self-reflection methods \cite{liang2023encouraging}. Although our objective differs, we similarly employ debate--- not to serve as a means to improve the final output but, rather, as the outcome itself--- using the tree directly to generate refined summaries of differences between research papers.

\begin{figure*}[ht!]
    \centering
    \includegraphics[width=1.0\textwidth]{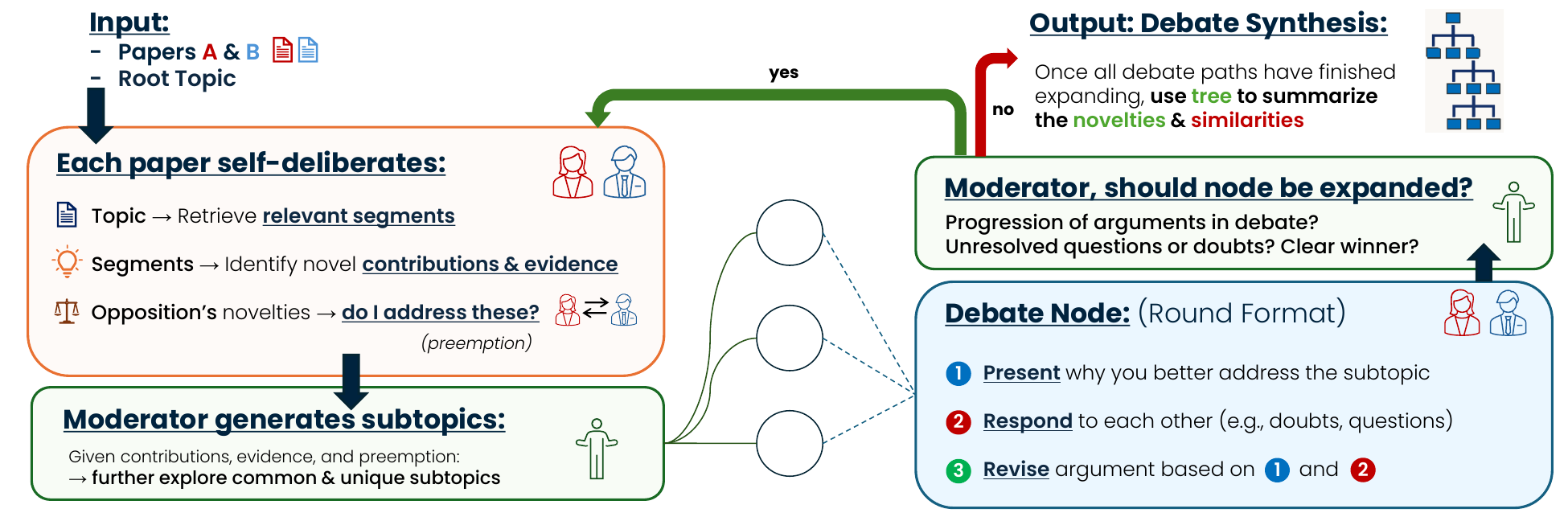}
    \caption{
    \label{fig:tree} We propose Tree-of-Debate, a novel framework which guides a multi-persona debate using a retrieval-augmented tree. $A \rightarrow B$ 
within the diagram translates to the statement, ``Given $A$, a persona arrives at $B$''.}
\end{figure*}

\subsection{Comparative Summarization} Generating comparative summaries is challenging due to the diverse ways that differences between two entities can be represented. Traditional graph-based methods \cite{comparative_graph, Strhle2023} classify sentences as a claim, similarity, or difference and score them to produce extractive summaries. While we use extractive summarization questions for self-deliberation, ultimately we aim to generate an \textit{abstractive} summary that synthesizes the debate results. More recent works \cite{explainingrelationships} fine-tune models to generate explanation sentences by first extracting in-text citation sentences that compare a principal document with a cited one, then maximizing the probability of generating the explanation given the two documents; however, this approach typically yields only a single sentence, which may not fully capture the nuanced differences between papers.

\subsection{Generation of Related Works Sections} 
\par Multi-document summarization consolidates information from various sources, a task that grows in importance as scientific literature expands \cite{comparative_graph}. Certain works within the HCI space \cite{palani2023relatedly,lee2024paperweaver}, which have designed off-the-shelf, interactive systems. However, from a methodological standpoint, one approach \cite{ur3wg} expands a paper's abstract into semantically similar sentences to form search queries for retrieving relevant papers, and then uses in-context examples to generate related work sections. On the other hand, DIR \cite{dir} employs a structured fine-tuning process by prompting a language model to extract commonalities and differences from candidate summaries compared to a gold standard. However, these methods face limitations: the former is highly dependent on the quality of its in-context example without clear guidelines on how to structure the related work. The latter requires fine-tuning the model for each dataset and relies on similarity matching that can restrict summary content \cite{contribution_sum, whats_new}. In contrast, our method uses debate rounds to guide the summary structure, operates at inference-time without training-- making it domain-agnostic-- and leverages the reasoning capabilities of language models to identify isomorphic properties of ideas beyond mere semantic similarity.

\section{Methodology}

\textsc{Tree-of-Debate} aims to determine and compare the fine-grained scientific claims of two papers through a methodology inspired by formal debate. Our overall framework is presented in Figure \ref{fig:tree}.
\subsection{Preliminaries}

\subsubsection{\textbf{Problem Formulation}} 
We assume two papers, $p_1$ and $p_2$, and a topic $n_0$ (e.g., ``inference-time LLM reasoning methods'') are provided as input by the user. Our goal is to determine the specific novelties, incremental additions, and equivalent contributions relevant to $n_0$ between $p_1$ and $p_2$, producing a \textbf{debate tree $T$} with a corresponding \textbf{comparative summary} $S$. In $T$, each node $n_i$ represents a topic (with $n_0$ as the root), where topic $n_i$ guides the specific debate occurring at that node. Topics may pertain to both papers or only one (e.g., in Figure \ref{fig:hierarchy}, only $p_i=B$ includes ``Backtracking''). An edge from $n_i$ to $n^i_j$ indicates that $n_j$ is a subtopic of $n_i$ that merits further exploration.

\subsubsection{\textbf{Segment-level Retrieval}}
\label{sec:retrieval}
An effective debate is contingent on an individual's preparation \textit{before} the debate and their ability to retrieve knowledge dynamically \textit{during} the debate. We employ a retrieval embedding model \cite{bge_embedding} and cosine-similarity to compute and rank segment-level embeddings. We chunk each paper into roughly three-sentence segments such that it is easily comprehensible during the debate.

\subsection{\textsc{Tree-of-Debate} Setup}

\par We conduct a multi-persona debate between two paper personas, $p_1$ and $p_2$, based on the high-level claim, \textit{\underline{$p_1$ is better than the $p_2$ for topic $n_i$}}. Our goal is not to determine a final debate ``winner'' but to capture the \textbf{specific reasoning} \textbf{\textit{induced}} by the debate format, reflected in the progression of arguments and the degree of novelty in each paper’s claims. While we explored novelty-specific claims (e.g., ``$p_1$’s contribution towards $n_i$ is more novel than $p_2$’s''), this led to more surface-level arguments. Moreover, a paper often features a breadth of claims/ideas, motivating a debate structure that is flexible to explore these different angles independently. Thus, we propose a \textbf{\textsc{Tree-of-Debate}} (ToD), $T$, where each node represents a round of debate (Section \ref{sec:debate_round_format}). A directed edge from parent node $n_p$ to child node $n^p_c$ indicates the debate progressing into one of $n_p$'s subtopics $n^p_c$ (out of potentially $k$ subtopics). Leaf nodes indicate no further progression in argumentation is evident.

\subsubsection{Constructing the Personas} We leverage an LLM agent to embody each debate persona, allowing for retrieved information from the papers and the debate history to be easily integrated into its context:
\begin{itemize}[leftmargin=*]
    \item \textbf{Papers:} Each paper persona is given the title, abstract, and retrieved segments relevant to the starting topic $n_0$ (updated at each self-deliberation stage (Section \ref{sec:self-deliberate})). Each paper persona's $p_i$ role is to argue that their contributions towards the topic $n_i$ are better than persona $p_j$'s.
    \item \textbf{Moderator:} Using the same underlying model as the paper personas, the moderator (i) identifies key debate subtopics for determining the papers' similarities and differences, (ii) judges the progression of the debate based on the authors' arguments, and (iii) synthesizes the debate tree into a comparative summary.
\end{itemize}

\subsubsection{\textbf{Tree Node Format}}
\label{sec:debate_round_format}
Each tree node with topic $n_i$ undergoes a three-stage debate (pre-, during, and post-debate). The format is as follows:
\begin{enumerate}[leftmargin=*]
    \item \textbf{Self-Deliberation (\textit{Section \ref{sec:self-deliberate}})}: Each persona $p_a$ retrieves segments $S^a_i$ relevant to $n_i$, generates $k$ claims $C^a_i$ for their novel contributions, cites corresponding evidence $E^a_i \subseteq S^a_i$, and collects counter-evidence from their own paper $\widetilde{E}^a_i$. The moderator then selects $k$ new subtopics for the next level of $k$ children, $n_j^i \in N^i$.
    
    \item \textbf{Debate (Prompts \ref{prompt_persona: present_arguments}, \ref{prompt_persona: respond_to_argument}, and \ref{prompt_persona: revise_argument})}: For each child debate node $n_j^i$, each persona $p_a$ (i) presents an argument that $p_a$ is \underline{better} than $p_b$ on $n_j$, (ii) responds to the opposing argument, and (iii) revises their argument accordingly.
    
    \item \textbf{Determine Expansion \textit{(Section \ref{sec:expansion})}}: Based on the debate at $n_j^i$, the moderator determines whether the arguments progressed or introduced any unresolved questions meriting another round. If so, the moderator triggers self-deliberation for $n_i \rightarrow n^i_j$.
\end{enumerate}

\subsection{Self-Deliberation}
\label{sec:self-deliberate}
\par Self-deliberation is an argumentative strategy \cite{tindale2020self} that enables one to ``argue with oneself'' by considering alternative views, aiming to arrive at the best, most well-justified conclusion. We integrate self-deliberation into our multi-persona debate for a given topic node $n_i$ and paper $p_{a\in\{1,2\}}$:
\begin{enumerate}[leftmargin=*]
    \itemsep=0em
    \item Retrieve relevant segments $S_a^i$ from $p_a$ that are closely related to $n_i$.
    \item Generate $k$ claims $c_j \in C^a_i$ on the \textit{novel contributions} of $p_a$ toward $n_i$. Each claim includes a title, description, and a set of mapped evidence $E^a_{(i, j)}\subseteq S_i^a$ (see Prompt \ref{prompt_persona: generate_arguments}).
    \item Preempt $p_b$'s contributions, where $C^b_i$ is exposed in-context to $p_a$ and $p_a$ retrieves another round of evidence $\widetilde{E}^a_i$ aimed at targeting $p_b$'s claims.
    \item The moderator then uses $E_i$, $\widetilde{E}_i$, and $C_i$ to generate a list of \textbf{subtopics} for further exploration.
\end{enumerate}

\par{\textbf{Retrieving relevant segments.}} For each paper $p_{a\in\{1,2\}}$, we retrieve the top $\delta$ segments $S^i_{a\in\{1,2\}}$ conditioned on node topic $n_i$ using the retrieval embedding model \cite{bge_embedding} (Section \ref{sec:retrieval}). We embed $n_i$ using the query format: ``[\texttt{topic name}] : [\texttt{topic description}]'' (the moderator generates a description for each non-root node). These segments form two separate pools of evidence per paper to compose their novelty claims.

\par{\textbf{Preemption.}} For each of paper $p_b$'s novelty claims $c^b_{(i,j)} \in C^b_i$ with its corresponding evidence $E^b_i$, we retrieve additional segments $\widetilde{E}^a_{(i,j)}$ from $p_a$ using the concatenated title and description of $c^b_{(i,j)}$ as the query. Each retrieved segment $e$ is then filtered using an LLM-based step (Prompt \ref{prompt_persona: is_irrelevant_evidences}) that evaluates whether $e$ (1) \underline{\textit{supports}}, (2) \underline{\textit{refutes}}, (3) \underline{\textit{clarifies}} $p_b$'s claim, or (4) is \underline{\textit{irrelevant}}. While redundant, we notice that explicitly including (4) as an option helps with filtration performance. If either (1-3) are true or (4) is false, then $e$ is filtered out. If $|\widetilde{E}^i_{(a, j)}| = 0$, we indicate that $p_a$ does not address $p_b$'s claim, $c^b_{(i,j)}$. Overall, preemption allows the paper personas to be better prepared for their opposition's arguments ahead of the debate.

\par{\textbf{Subtopic Generation.}} Using each paper's title, abstract, $C_i$, $E_i$, and $\widetilde{E}_i$, the moderator generates $k$ \textbf{subtopics} $n^i_j \in N^i$ that should be further explored (Prompt \ref{prompt_mod: generate_topics}). The moderator maps each subtopic to at least one claim $c_i$ from either $p_1$ and/or $p_2$, forming child debate nodes that explore overlapping topics (e.g., ``Reasoning Evaluation'' in Figure \ref{fig:hierarchy}) or topics potentially unique to one paper (e.g., ``Multiple Paths'').

\subsection{Debate Tree Expansion \& Synthesis}
\label{sec:expansion}

\par While we motivate the personas to examine and debate whether their work proposes a \textit{better} idea than their opposition, this mechanism is intended to (1) emphasize the \textbf{\textit{reasoning behind the idea}} and (2) provoke debate on the \textbf{\textit{novelty behind the ideas}}, relative to each other. In other words, we hypothesize that two very similar ideas (e.g., ``Reasoning Evaluation'' in Figure \ref{fig:hierarchy}) will typically lead to a \textit{longer} debate subtree on which approach is better. Conversely, a uniquely novel approach or task proposed in $p_1$, relative to $p_2$, may result in a \textit{shorter} debate as the moderator will ideally determine that $p_2$ does not address $p_1$'s claim in their work and thus has a weak argument (e.g., ``Backtracking'' in Figure \ref{fig:hierarchy}). To facilitate this process, the moderator's core tasks are detailed below.

\subsubsection{\textbf{Determining Round Depth Expansion}}
\par For debate node $n_i$, the moderator assesses the following (Prompt \ref{prompt_mod: is_expand}):
\begin{enumerate}[leftmargin=*]
    \item \textbf{Argument Progression}: Is there sufficient evolution in the arguments or new, deeper concepts being introduced to justify further debate?
    \item \textbf{Meaningful Questions}: Have clarifying questions been raised that remain unanswered and merit further discussion? If no questions are raised, the moderator returns False.
    \item \textbf{Clear Winner}: Is it clear that one paper has won the debate, as their contributions are truly better and do not warrant deconstruction (to determine which \textit{subcomponents} are truly better)?
\end{enumerate}
If either (1) or (2) holds true, or if (3) does not indicate a clear winner, the moderator proceeds to the self-deliberation stage (Section \ref{sec:self-deliberate}) to identify new subtopics and expand $n_i$ to $n^i_j \in N^i$. Otherwise, expansion stops for that debate path. A maximum tree depth is also enforced.

\subsubsection{\textbf{Debate Synthesis}}
\par Once ToD has converged (i.e., all debate paths have been adequately expanded), the moderator synthesizes the entire debate tree into a paragraph-long comparative summary. The debate tree is provided in-context, with each node $n_i \in T$ containing the following information: node topic \textbf{title}, node topic \textbf{description}, persona $p_1$'s \textbf{revised argument} (at the end of the debate round), and persona $p_2$'s \textbf{revised argument}. The synthesis should first explain the papers' novelty similarities and then detail their differences, with greater emphasis on the latter. We provide the prompt in Prompt \ref{prompt_mod: summarize_debate} and an example subtree in Appendix \ref{sec:tree_example}.

\section{Experimental Design}
\par We choose \texttt{Llama-3.1-Nemotron-70B-Instruct-\\HF}, an \textit{open-source model}, as the base model for all experiments. We sample from the top 1\% of the tokens and use the same temperature settings across all samples (details on setting and hardware provided in Appendix \ref{appendix: settings}; we also conduct supplementary studies on replacing our base LLM and retrieval model in Appendices \ref{appendix:retrieval} and \ref{appendix:base_llm}). Finally, all prompts are provided in Appendix \ref{appendix: prompts}.

\subsection{Dataset}
\label{sec:dataset}
\par No datasets currently exist for comparing \textit{non-citing} pairs of scientific papers--- an overlooked setting, especially given the growing scale of literature where not all relevant work can be cited. Consequently, we aimed to construct a dataset with papers that both cite and do not cite each other, in order to test the robustness of Tree-of-Debate. However, novelty comparison between papers is a highly specialized and expensive task, requiring rich domain expertise to verify-- especially if such papers do not explicitly cite one another. Thus, we gathered five domain expert researchers (details provided in Appendix \ref{appendix: annotators}) to construct a dataset of 100 paper pairs across natural language processing, data mining, electrical engineering, and aerospace engineering (further details provided in Appendix \ref{appendix:dataset}). Each researcher identified at least five papers they were highly familiar with, such that they could perform a detailed and informed human evaluation. They were instructed to annotate each pair with a root topic and whether: (1) the papers roughly focus on the same task but differ in \textbf{methodology}, or (2) they work on different \textbf{tasks} that are applied to \textit{similar} motivations. Furthermore, they noted if the papers explicitly \textbf{cited} each other \textbf{or not}. Table \ref{tab:dataset} shows the dataset distribution.

\begin{table}[h!]
    \footnotesize
    \centering
    \begin{tabular}{l|cc|c}
        \toprule
         \textbf{Category} & \textbf{Method} & \textbf{Task} & \textbf{Total}\\
         \midrule
         \textbf{Cited} & 15 & 15 & 30 \\
         \textbf{Not Cited} & 30 & 40 & 70 \\
         \midrule
         \textbf{Total} & 45 & 55 & 100 \\
         \bottomrule
    \end{tabular}
    \caption{\# of paper pairs/summaries per category.}
    \label{tab:dataset}
\end{table}

\subsection{Baselines}
\label{sec:baselines}
\par Given that our primary goal is to demonstrate the difference in inference-time comparative reasoning capabilities between ToD and current LLMs, we design the following prompting-based baselines: (1) \textbf{Single Stage} uses the title, abstract and introduction sections of both papers to directly generate a comparative summary \cite{MartinBoyle2024ShallowSO}; (2) \textbf{Two Stage} first individually summarizes each paper based on the title, abstract and introductions, and then uses the generated summaries to generate a comparative summary \cite{Zhang2024FromCT}.

\par To contextualize improvements from each component in ToD, we construct the following ablative methods: (1) \textbf{ToD (No Tree)} removes the tree structure by merging child arguments into one and considering the combined subtopic as the debate node topic; (2) \textbf{ToD (No SD)} removes self-deliberation (SD) to test the impact of iterative retrieval based on debate progression. No SD relies on the title, abstract, and introduction of each paper instead of retrieving based on the subtopic. We use the same LLM for all comparisons. Complete baseline and ablation details are provided in Appendix \ref{appendix:baselines}.

\subsection{Evaluation Metrics}
\par The same domain-experts from Section \ref{sec:dataset} manually evaluate each of their chosen pairs in-depth, assessing various qualities of the 100 summaries. We normalize each of the scores below and scale them by 100 for the final results in Table \ref{tab:method_results}:
\begin{itemize}[leftmargin=*]
    \itemsep-0.3em
    \item \textbf{Factuality}: \textit{How factual is the summary?} Each sentence is given a 1/0 binary score for factuality, and the scores are averaged across the summary.
    \item \textbf{Breadth}: \textit{Is the summary comprehensive and complete?} Each summary is rated from 0-4 (``\textit{not at all}'' to ``\textit{very}'').
    \item \textbf{Contextualization}: 
    \textit{Does the summary explain and/or justify the posed similarities/differences between the papers, as opposed to just mentioning them?} Each summary is rated from 0-4 (``\textit{not at all}'' to ``\textit{very}'').
\end{itemize}

\par We note that we strive to balance the need for ``objective'' metrics with some added flexibility, necessary for this challenging evaluation task. Thus, as we are exclusively using manual domain-expert evaluation for all of our metrics, we include scoring guidelines like ``the summary covers all major points, but still is not what I would expect'' (criteria for breadth score of 3). We provide the full metric guidelines in Appendix \ref{appendix:eval_metrics}, in addition to an annotation/evaluation agreement study in Appendix \ref{appendix:agreeement}.

% First Table: Method + Cite and Method + Not Cited
\begin{table*}[ht]
\small
\centering
\renewcommand{\arraystretch}{1.1}
\caption{We showcase \textsc{Tree-of-Debate}'s strong performance across all comparison settings. \textbf{Bolded} values indicate the top score; \underline{underlined} indicates second-highest. We include two ablations which remove the tree structure (\textit{No Tree}) and self-deliberation (\textit{No SD}), respectively.}
\label{tab:method_results}
\resizebox{\textwidth}{!}{%
\begin{tabular}{|c|ccc|ccc|ccc|}
\hline
\multirow{2}{*}{\textbf{Baseline}} 
& \multicolumn{3}{c|}{\textbf{Method + Cite}} 
& \multicolumn{3}{c|}{\textbf{Method + Not Cited}} 
& \multicolumn{3}{c|}{\textbf{Overall (Method)}} \\ \cline{2-10}
& \textit{Breadth} & \textit{Context} & \textit{Factuality} 
& \textit{Breadth} & \textit{Context} & \textit{Factuality} 
& \textit{Breadth} & \textit{Context} & \textit{Factuality} \\ \hline

% Single Stage
Single Stage   
& \underline{93.33} & \underline{80.00} & \textbf{93.48} 
& \underline{89.81} & 73.15 & \underline{94.90} 
& \underline{91.07} & 75.59 & \underline{94.39} \\

% Two Stage
Two Stage      
& 85.00 & 71.67 & 91.42 
& 87.03 & 72.22 & 94.74 
& 86.31 & 72.02 & 93.69 \\ \hline

% ToD (No Tree)
\textit{ToD (No Tree)}  
& 81.67 & 63.33 & 91.25 
& 79.63 & 70.37 & 89.46 
& 80.36 & 67.86 & 89.82 \\

% ToD (No SD)
\textit{ToD (No SD) }
& 91.67 & 76.67 & 82.33 
& 86.11 & \textbf{94.44} & 76.33 
& 88.10 & \underline{88.10} & 78.21 \\

% \textbf{Tree-of-Debate} 
\textbf{Tree-of-Debate}       
& \textbf{96.67} & \textbf{93.33} & \underline{92.80} 
& \textbf{93.52} & \underline{93.52} & \textbf{96.80} 
& \textbf{94.64} & \textbf{93.45} & \textbf{95.37} \\ \hline

\end{tabular}%
}
\newline\newline\newline
\resizebox{\textwidth}{!}{%
\begin{tabular}{|c|ccc|ccc|ccc|}
\hline
\multirow{2}{*}{\textbf{Baseline}} 
& \multicolumn{3}{c|}{\textbf{Task + Cite}} 
& \multicolumn{3}{c|}{\textbf{Task + Not Cited}} 
& \multicolumn{3}{c|}{\textbf{Overall (Task)}} \\ \cline{2-10}
& \textit{Breadth} & \textit{Context} & \textit{Factuality} 
& \textit{Breadth} & \textit{Context} & \textit{Factuality} 
& \textit{Breadth} & \textit{Context} & \textit{Factuality} \\ \hline

% Single Stage
Single Stage   
& 81.82 & 65.91 & \underline{94.44} 
& \underline{88.46} & 71.15 & 92.31 
& 87.00 & 70.00 & \underline{92.78} \\

% Two Stage
Two Stage      
& 89.29 & 82.14 & \textbf{97.14} 
& 86.84 & 78.29 & \textbf{94.18} 
& \underline{87.22} & 78.89 & \textbf{94.54} \\ \hline

% ToD (No Tree)
\textit{ToD (No Tree)}  
& 67.86 & 67.86 & 61.11 
& 83.55 & 73.03 & 91.43 
& 81.11 & 72.22 & 89.74 \\

% ToD (No SD)
\textit{ToD (No SD)} 
& \textbf{96.43} & \underline{96.43} & 75.00 
& 81.58 & \underline{94.08} & 82.27 
& 83.89 & \underline{94.44} & 81.79 \\

% \textbf{Tree-of-Debate} 
\textbf{Tree-of-Debate}       
& \underline{92.86} & \textbf{100} & 87.89 
& \textbf{96.05} & \textbf{96.05} & \underline{94.02} 
& \textbf{95.19} & \textbf{97.11} & 92.36 \\ \hline

\end{tabular}%
}
\end{table*}

\begin{table*}[ht!]
    \small
    \centering
    \renewcommand{\arraystretch}{1.5}
    \caption{Debate Round for Tree of Thoughts (ToT) vs. Chain-of-Thought (CoT); turns are summarized and color-coded based on strength of argument (red/weak $\rightarrow$ yellow/medium $\rightarrow$ green/strong).}
    \begin{tabular}{|p{1.1cm}|p{14cm}|}
        \hline
        \textbf{Stage} & \textbf{Dialogue for Subtopic: 	``Generalizability and Flexibility of Prompting Methods''} \\ 
        \hline
        \textbf{Present} & 
        \begin{tabular}[t]{@{}p{14cm}@{}}
            \cellcolor{Red!5}\textbf{ToT}: ToT \textit{enhances flexibility} via deliberate reasoning by exploring multiple paths, self-evaluation of choices, and lookahead/backtracking capabilities-- shown on a diverse set of tasks like Game of 24 and Creative Writing. This is in contrast to CoT's lack of deliberate reasoning and adaptability. \\ \hline
            \cellcolor{Green!5}\textbf{CoT}: CoT excels in \textit{generalizability through its simplicity}, requiring minimal examples to achieve strong results across tasks. Unlike ToT, it \textit{avoids complex setups and can generalize to unseen tasks}, making it more adaptable with fewer computational demands.
        \end{tabular} \\ 
        \hline
        \textbf{Respond} & 
        \begin{tabular}[t]{@{}p{14cm}@{}}
            \cellcolor{Yellow!5}\textbf{ToT}: While CoT is simpler, ToT’s added complexity is a necessary \textit{trade-off for deeper reasoning tasks}, requiring non-trivial planning and search. CoT’s ability to handle deeply nested or conditional reasoning remains unproven. \\ \hline
            \cellcolor{Green!5}\textbf{CoT}: ToT may help in specific tasks, but CoT's simplicity deliberately ensures \textit{broader applicability} without overfitting to specific task requirements. Though direct evidence for nested reasoning is lacking, CoT’s state-of-the-art GSM8K performance suggests complex reasoning robustness. A future study can compare both methods on nested reasoning to settle this question empirically.
        \end{tabular} \\ 
        \hline
        \textbf{Revise} & 
        \begin{tabular}[t]{@{}p{14cm}@{}}
            \cellcolor{Green!5}\textbf{ToT}: ToT has \textit{enhanced flexibility via adaptive reasoning path exploration}-- exploring multiple reasoning paths and leveraging self-evaluation and lookahead/backtracking to tackle complex tasks-- as shown on our diverse problem domains. Our deliberate decision-making is advantageous for tasks with \textit{deeply nested or highly conditional reasoning paths}, where CoT may falter. \\ \hline
            \cellcolor{Green!5}\textbf{CoT}: CoT \textit{maintains strong generalizability by balancing simplicity and effectiveness}-- including those with complex reasoning requirements-- \textit{without overfitting} to specific task needs. This is demonstrated through its SoTA performance on GSM8K.
        \end{tabular} \\ 
        \hline
    \end{tabular}
    \label{tab:evolution}
\end{table*}

\section{Experimental Results}

\par{\textbf{Overall Performance \& Analysis.}} Table \ref{tab:method_results} shows the performance of Tree-of-Debate (ToD) compared with the baselines on factuality, breadth of comparison (completeness), and contextualization. We observe that the domain-experts found ToD summaries $6.85\%$ more complete and $25.98\%$ more contextualized compared to the most competitive baseline. This observation indicates that multi-persona debate trees help analyze pairs of papers to uncover more fine-grained contributions, as well as identifying \textit{connections} between the papers. Given that all samples were carefully annotated and evaluated by domain-experts, we are able to draw several interesting insights, which we list below:

\begin{table*}[h!]
    \centering
    \small
    \renewcommand{\arraystretch}{1.1}
    \begin{tabular}{|p{0.75cm}|p{14.5cm}|}
        \hline
        \textbf{Model} & \textbf{Comparative Summary Excerpts} \\ 
        \hline
        \textbf{Two Stage} & MEGClass focuses on mutually-enhancing text granularities, iteratively refining its model through feedback between document, sentence, and word levels, and \textit{\textcolor{red}{requires an initial set of weakly labeled data}}. In contrast, LOTClass \textit{\textcolor{red}{innovates by relying solely on label names}} to train its model, employing a language model self-training approach that \textit{\textcolor{red}{obviates the need for any labeled documents}}, achieving high accuracy without human-labeled data. \\ 
        \hline
        \textbf{ToD} & \textit{\textcolor{ForestGreen}{Both papers leverage limited labeled data}} to achieve robust performance. Both MEGClass and LOTClass also employ \textit{\textcolor{ForestGreen}{iterative refinement techniques}}, with MEGClass using iterative feedback and LOTClass utilizing self-training on unlabeled data. MEGClass \textit{\textcolor{ForestGreen}{uniquely leverages mutually-enhancing text granularities}}, capturing both coarse- and fine-grained context signals to provide a more comprehensive understanding of class labels. In contrast, LOTClass relies \textit{\textcolor{ForestGreen}{solely on label names}}, leveraging pre-trained language models to drive a self-training mechanism. Furthermore, MEGClass's adaptive granularity balancing approach provides robustness to label name selection, whereas LOTClass's \textit{\textcolor{ForestGreen}{reliance on label names alone may introduce biases.}} \\ 
        \hline
    \end{tabular}
    \caption{Case study on two weakly supervised text classification works \cite{kargupta-etal-2023-megclass,meng-etal-2020-text}.}
    \label{tab:summary}
\end{table*}

\par{\textbf{\underline{Structured} debate improves the \underline{contextualization} of comparative summaries.}} Our results show that \textsc{Tree-of-Debate} significantly improves contextualization, achieving an average score of $95.21\%$ across all settings, compared to $75.57\%$ for the strongest baseline (Two-Stage). Our domain-expert evaluators frequently observed that the LLM-generated summaries (Single and Two Stage) at face value mention a breadth of \textit{specific} contributions made by each paper, noting them as either similarities or differences. However, these tend to resemble \textit{\textbf{extractive}} summaries, where phrases that are \textit{semantically similar or dissimilar} are extracted from the papers' abstracts and introduction and are posed as similarities and differences respectively, with no context provided on \textit{why} this is the case. We further analyze this finding in Section \ref{sec:summary_case_study}. Moreover, we see that without the tree-structure, the debate's analysis quality deteriorates. Specifically, for our ``No Tree'' ablation study (Table \ref{tab:method_results}), we modify our method to combine all proposed subtopics by the moderator (Section \ref{sec:self-deliberate}) into a single high-level topic. This leads to lower contextualization and breadth of contributions discussed, due to reasoning difficulty in disentangling the contributions during the debate. Thus, \textit{\underline{structuring} the debate is necessary to experience its contextualization benefits}.

\begin{table*}[h!]
\scriptsize
\centering
\begin{tabular}{lccc|ccc|ccc|ccc}
\toprule
\textbf{Model} 
& \multicolumn{3}{c|}{\textbf{NLP}} 
& \multicolumn{3}{c|}{\textbf{Data Mining}} 
& \multicolumn{3}{c|}{\textbf{Electrical}} 
& \multicolumn{3}{c}{\textbf{Aerospace}} \\
\cmidrule(lr){2-4} \cmidrule(lr){5-7} \cmidrule(lr){8-10} \cmidrule(lr){11-13}
& \textbf{Breadth} & \textbf{Context} & \textbf{Fact}
& \textbf{Breadth} & \textbf{Context} & \textbf{Fact}
& \textbf{Breadth} & \textbf{Context} & \textbf{Fact}
& \textbf{Breadth} & \textbf{Context} & \textbf{Fact} \\
\midrule
Single 
& 86.74$^\dagger$ & 72.35 & 93.28
& 85.00$^\dagger$ & 65.00 & 90.13$^\dagger$
& \textbf{100.00} & 80.77$^\dagger$  & 93.80$^\dagger$ 
& 85.00$^\dagger$  & 85.00$^\dagger$ & 95.28$^\dagger$ \\
Two 
& 84.85 & 74.24$^\dagger$ & 94.13$^\dagger$
& 70.00 & 71.67$^\dagger$ & 84.68
& \textbf{100.00} & 80.77$^\dagger$  & \textbf{94.14}
& 85.00$^\dagger$  & \textbf{90.00} & \textbf{97.14} \\
\midrule
ToD 
& \textbf{95.08} & \textbf{95.83} & \textbf{94.78}
& \textbf{91.67} & \textbf{100.00} & \textbf{90.99}
& 98.08$^\dagger$ & \textbf{88.46}& 91.36
& \textbf{100.00} & \textbf{90.00} & 88.64 \\
\bottomrule
\end{tabular}
\caption{Domain-level performance for each model. Bold values indicate best in column; $^\dagger$ marks second-best.}
\label{tab:domain_metrics}
\end{table*}

\par{\textbf{\underline{Retrieval-augmented} debate ensures factuality and breadth of comparison.}} In Table \ref{tab:method_results}, we compare the performance between ToD and ``No SD'', where the latter ablation replaces the iterative retrieval step with simply providing the paper's title, abstract, and introduction as in-context evidence across the entire debate tree (same setting as the two baselines). We observe that No SD still experiences the strong benefits of the structured debate, evident through its similarly high contextualization score (on average, $91.59\%$, compared to ToD's $95.46\%$ and Two Stage's $75.80\%$). We note a drop in the factuality and breadth of the summary. The information present in the abstract and introduction is not detailed or deep enough to facilitate fine-grained discussion of the paper. This leads to more noise, and hallucinations, which negatively impacts No SD's breadth and factuality scores. Therefore, \textit{\underline{iterative evidence retrieval} is necessary for exploring in breadth and minimizing hallucinations}.

\par{\textbf{\textsc{ToD} is \underline{robust} to all comparison settings.}} We analyze four different comparison settings (as detailed in Section \ref{sec:dataset}), where two papers may or may not cite each other and differ in either their \textit{task} or \textit{method}. We can see through Table \ref{tab:method_results} that regardless of their comparison type, ToD demonstrates consistently high performance-- with only an average standard deviation $\sigma=2.49$ across breadth and soundness, while Single Stage's $\sigma = 5.32$ and Two Stage's $\sigma=3.39$. While there are no consistent trends across the different settings, studying them allows us to \textit{ensure the \underline{consistency} of ToD}.

\subsection{Qualitative Case Study}
\label{sec:summary_case_study}
\par{\textbf{Evolution via Critical Reasoning.}} Our approach enables paper personas to refine their comparisons by addressing counterarguments elicited through debate. Table \ref{tab:evolution} illustrates this through a debate round between the Tree of Thoughts (ToT) and Chain-of-Thought (CoT) papers on their generalizability and flexibility ($n_i$). ToT initially highlights its \textit{flexibility} through deliberate reasoning, exploring multiple paths with lookahead/backtracking. However, CoT counters that ToT's complexity hinders broad applicability, whereas CoT can easily generalize to a multitude of tasks with minimal setup while still achieving state-of-the-art results. In the revision stage, both refine their arguments: ToT emphasizes adaptive exploration for \textit{complex reasoning}, while CoT maintains its simplicity, now citing empirical success in complex tasks. It is interesting to note that during the \textit{Respond} stage, CoT \textbf{\textit{suggests a future study}} that would help validate the claims made in the debate.

\par{\textbf{Contextualized Summaries.}} The baselines' summaries tend to be extractive, not explaining why certain comparisons are important and also mistaking similarities for differences if their wording is semantically dissimilar. Table \ref{tab:summary} demonstrates this finding; Two Stage \textit{mistakenly states} that MEGClass ``requires an initial set of weakly labeled data'' while LOTClass ``innovates by relying solely on label names''. However, this is false-- both methods only require the class label names. This similarity is identified by ToD, which uses it to contextualize the methods' differences: both use label names, however MEGClass considers all three levels of text granularity, while LOTClass solely relies on the label names. ToD also provides \textit{further insight} into LOTClass's over-reliance on label names potentially introducing biases. ToD's debate format elicits critical reasoning and allows for a more insightful, contextualized comparison. Appendices \ref{sec:tree_example} and \ref{appendix:summary} include in-depth qualitative analyses on an additional sample's tree and output summaries.

\subsection{Domain-Level Performance}
\par We study Tree-of-Debate's performance in each specific domain, with the quantitative results shown in Table \ref{tab:domain_metrics}. We can see that overall, Tree-of-Debate (1) has the biggest overall gains within the NLP \& Data Mining domains, (2) features large contextualization gains in electrical engineering, and (3) shows both context and breadth gains for aerospace. However, we do note that the factuality score does drop for the latter two domains. We hypothesize that this is due to the relatively smaller scale of research-level information of these fields within the model’s pre-training knowledge, and thus increasing the difficulty of the debate-specific reasoning.

\subsection{Influence of Debate Depth}

\par We include a qualitative case study in Appendix \ref{appendix: depth} on how the summaries evolve as we vary the depth from 1 to 3. \textbf{As depth increases, both factuality and contextualization generally increase.} Based on the expert’s feedback, it seems that depth=1 produces shallow summaries that lean more on their surface-level knowledge of the paper’s abstract and high-level points--- leading to less contextualized summaries. This also impacts the factuality, leading to more superfluous/overly positive statements. However, this significantly improves as depth increases, where the \textit{personas are able to dive deeper into fine-grained points} and are able to further \textit{refine/tone-down their exaggerated arguments}.

\subsection{Contextualization \& Factuality Trade-off}
\par We observe a \textbf{\textit{slight trade-off between contextualization and factuality}}. When factuality mistakes do occur, they are primarily made when facts are mixed up between the papers when summarizing their debate. For example, when comparing \citet{sener2017active} and \citet{kumari-etal-2024-end}, ToD thinks \citet{sener2017active} is ``\textit{applicable across various domains, including image classification and ICL [in-context learning]}'' for LLMs. In fact, \citet{sener2017active} is for image classification, while \citet{kumari-etal-2024-end} is for ICL.

\par Furthermore, especially in more challenging comparisons where the overlap between the papers may not be explicit (e.g., task-based comparisons), ToD must take more of a ``leap'' with respect to its reasoning to identify similarities and differences between the papers within the debate. Overall, this more challenging setting leads to a drop in factuality compared to baselines. We provide a more comprehensive error analysis in Appendix \ref{appendix:error_analysis}.

\section{Conclusion}
Automatic summarization is essential for managing the growing volume of research. We introduce \textbf{\textsc{Tree-of-Debate}}, a structured approach that models papers as personas engaging in a debate to extract their key similarities and differences. Our method organizes debates into a hierarchical structure to produce abstractive and contextualized summaries while preserving their factuality. With thorough domain-expert empirical evaluation and qualitative case studies, we demonstrate that Tree-of-Debate significantly outperforms baselines.

\section{Limitations \& Future Work}
\par We explore some limitations of our work. Within each debate round (each persona presenting their arguments, responding to one another, and revising their arguments), we note that a crucial element to a productive debate round is \textbf{\textit{each persona providing meaningful feedback}} (e.g., doubts, clarifying questions)  within the ``respond'' stage. However, the quality of this critical response may vary based on the difficulty of the task (e.g., a more fine-grained topic that has no presence within the model's existing pre-training dataset) and/or the size of the model.
\par Furthermore, we notice that as a debate path progresses to deeper levels, if certain evidence is not present to support a paper's  fine-grained claims, the personas begin to suggest potential future studies or even new methods (e.g., combining certain strengths of the two methods). While this does introduce some noise (as we can see through our competitive factuality scores, this is minimal) to our summarization process, these types of ``hallucinations'' in the debate present exciting new paths for research to explore.
\par Tree-of-Debate can be \textbf{\textit{extended to other general, complex reasoning tasks}} which can exploit our tree-based decomposition and debate-based critical feedback. For instance, complex quantitative reasoning problems can often be decomposed into several sub-problems, which can be represented within our tree structure. Each persona can instead represent a different approach to solve that specific sub-problem. We can also consider extending this to a negotiation setting, where there are various aspects to consider when determining the optimal compromise. For example, two parties can negotiate on a price of a car with respect to its make, model, mileage, etc. Each of these can be explored within their own respective subtrees.

\section{Acknowledgements}
\par This work was supported by the National Science Foundation Graduate Research Fellowship. Any opinions, findings,
and conclusions or recommendations expressed in this material are those of the author(s) and do not necessarily reflect the views of the National Science Foundation. This research also used the DeltaAI advanced computing and data resource, which is supported by the National Science Foundation (award OAC 2320345) and the State of Illinois. DeltaAI is a joint effort of the University of Illinois Urbana-Champaign and its National Center for Supercomputing Applications. We would like to also thank Mihir Kavishwar, Alexandra Leeming, and Shivam Agarwal for their help and feedback on our work.

\bibliography{custom}

\begin{thebibliography}{35}
\providecommand{\natexlab}[1]{#1}

\bibitem[{arXiv(2024)}]{arxiv_monthly_submissions}
arXiv. 2024.
\newblock arxiv monthly submission statistics.
\newblock \url{https://arxiv.org/stats/monthly_submissions}.
\newblock Accessed: 2024-10-13.

\bibitem[{Chan et~al.(2023)Chan, Chen, Su, Yu, Xue, Zhang, Fu, and Liu}]{chan2023chateval}
Chi-Min Chan, Weize Chen, Yusheng Su, Jianxuan Yu, Wei Xue, Shanghang Zhang, Jie Fu, and Zhiyuan Liu. 2023.
\newblock Chateval: Towards better llm-based evaluators through multi-agent debate.
\newblock \emph{arXiv preprint arXiv:2308.07201}.

\bibitem[{Chen et~al.(2022)Chen, Cai, Jiang, and Chen}]{comparative_graph}
Jingqiang Chen, Chaoxiang Cai, Xiaorui Jiang, and Kejia Chen. 2022.
\newblock \href {https://aclanthology.org/2022.coling-1.522} {Comparative graph-based summarization of scientific papers guided by comparative citations}.
\newblock In \emph{Proceedings of the 29th International Conference on Computational Linguistics}, pages 5978--5988, Gyeongju, Republic of Korea. International Committee on Computational Linguistics.

\bibitem[{Devlin et~al.(2019)Devlin, Chang, Lee, and Toutanova}]{devlin-etal-2019-bert}
Jacob Devlin, Ming-Wei Chang, Kenton Lee, and Kristina Toutanova. 2019.
\newblock \href {https://doi.org/10.18653/v1/N19-1423} {{BERT}: Pre-training of deep bidirectional transformers for language understanding}.
\newblock In \emph{Proceedings of the 2019 Conference of the North {A}merican Chapter of the Association for Computational Linguistics: Human Language Technologies, Volume 1 (Long and Short Papers)}, pages 4171--4186, Minneapolis, Minnesota. Association for Computational Linguistics.

\bibitem[{Fu et~al.(2023)Fu, Peng, Khot, and Lapata}]{self-play}
Yao Fu, Hao Peng, Tushar Khot, and Mirella Lapata. 2023.
\newblock Improving language model negotiation with self-play and in-context learning from ai feedback.
\newblock \emph{arXiv preprint arXiv:2305.10142}.

\bibitem[{Gunel et~al.(2024)Gunel, Wendt, Xie, Zhou, Vo, Fisher, and Tata}]{gunel2024strum}
Beliz Gunel, James~B Wendt, Jing Xie, Yichao Zhou, Nguyen Vo, Zachary Fisher, and Sandeep Tata. 2024.
\newblock Strum-llm: Attributed and structured contrastive summarization.
\newblock \emph{arXiv preprint arXiv:2403.19710}.

\bibitem[{Hayashi et~al.(2023)Hayashi, Kryscinski, McCann, Rajani, and Xiong}]{whats_new}
Hiroaki Hayashi, Wojciech Kryscinski, Bryan McCann, Nazneen Rajani, and Caiming Xiong. 2023.
\newblock \href {https://doi.org/10.18653/v1/2023.eacl-main.72} {What{'}s new? summarizing contributions in scientific literature}.
\newblock In \emph{Proceedings of the 17th Conference of the European Chapter of the Association for Computational Linguistics}, pages 1019--1031, Dubrovnik, Croatia. Association for Computational Linguistics.

\bibitem[{Kargupta et~al.(2024)Kargupta, Agarwal, Hakkani-Tur, and Han}]{kargupta2024instructassistllmbasedmultiturn}
Priyanka Kargupta, Ishika Agarwal, Dilek Hakkani-Tur, and Jiawei Han. 2024.
\newblock \href {https://arxiv.org/abs/2406.11709} {Instruct, not assist: Llm-based multi-turn planning and hierarchical questioning for socratic code debugging}.
\newblock \emph{Preprint}, arXiv:2406.11709.

\bibitem[{Kargupta et~al.(2023)Kargupta, Komarlu, Yoon, Wang, and Han}]{kargupta-etal-2023-megclass}
Priyanka Kargupta, Tanay Komarlu, Susik Yoon, Xuan Wang, and Jiawei Han. 2023.
\newblock \href {https://doi.org/10.18653/v1/2023.findings-emnlp.708} {{MEGC}lass: Extremely weakly supervised text classification via mutually-enhancing text granularities}.
\newblock In \emph{Findings of the Association for Computational Linguistics: EMNLP 2023}, pages 10543--10558, Singapore. Association for Computational Linguistics.

\bibitem[{Kumari et~al.(2024)Kumari, Wang, Das, Zhou, and Bilmes}]{kumari-etal-2024-end}
Lilly Kumari, Shengjie Wang, Arnav Das, Tianyi Zhou, and Jeff Bilmes. 2024.
\newblock \href {https://doi.org/10.18653/v1/2024.findings-naacl.209} {An end-to-end submodular framework for data-efficient in-context learning}.
\newblock In \emph{Findings of the Association for Computational Linguistics: NAACL 2024}, pages 3293--3308, Mexico City, Mexico. Association for Computational Linguistics.

\bibitem[{Kwon et~al.(2023)Kwon, Li, Zhuang, Sheng, Zheng, Yu, Gonzalez, Zhang, and Stoica}]{kwon2023efficient}
Woosuk Kwon, Zhuohan Li, Siyuan Zhuang, Ying Sheng, Lianmin Zheng, Cody~Hao Yu, Joseph~E. Gonzalez, Hao Zhang, and Ion Stoica. 2023.
\newblock Efficient memory management for large language model serving with pagedattention.
\newblock In \emph{Proceedings of the ACM SIGOPS 29th Symposium on Operating Systems Principles}.

\bibitem[{Lee et~al.(2024)Lee, Kang, Latzke, Kim, Bragg, Chang, and Siangliulue}]{lee2024paperweaver}
Yoonjoo Lee, Hyeonsu~B Kang, Matt Latzke, Juho Kim, Jonathan Bragg, Joseph~Chee Chang, and Pao Siangliulue. 2024.
\newblock \href {https://doi.org/10.1145/3613904.3642196} {Paperweaver: Enriching topical paper alerts by contextualizing recommended papers with user-collected papers}.
\newblock In \emph{Proceedings of the 2024 CHI Conference on Human Factors in Computing Systems}, CHI '24, New York, NY, USA. Association for Computing Machinery.

\bibitem[{Lerman and McDonald(2009)}]{lerman2009contrastive}
Kevin Lerman and Ryan McDonald. 2009.
\newblock Contrastive summarization: an experiment with consumer reviews.
\newblock In \emph{Proceedings of human language technologies: The 2009 annual conference of the North American chapter of the association for computational linguistics, companion volume: Short papers}, pages 113--116.

\bibitem[{Li et~al.(2024)Li, Wang, Zheng, and Zhang}]{li-etal-2024-loogle}
Jiaqi Li, Mengmeng Wang, Zilong Zheng, and Muhan Zhang. 2024.
\newblock \href {https://doi.org/10.18653/v1/2024.acl-long.859} {{L}oo{GLE}: Can long-context language models understand long contexts?}
\newblock In \emph{Proceedings of the 62nd Annual Meeting of the Association for Computational Linguistics (Volume 1: Long Papers)}, pages 16304--16333, Bangkok, Thailand. Association for Computational Linguistics.

\bibitem[{Liang et~al.(2024)Liang, He, Jiao, Wang, Wang, Wang, Yang, Shi, and Tu}]{liang2023encouraging}
Tian Liang, Zhiwei He, Wenxiang Jiao, Xing Wang, Yan Wang, Rui Wang, Yujiu Yang, Shuming Shi, and Zhaopeng Tu. 2024.
\newblock \href {https://doi.org/10.18653/v1/2024.emnlp-main.992} {Encouraging divergent thinking in large language models through multi-agent debate}.
\newblock In \emph{Proceedings of the 2024 Conference on Empirical Methods in Natural Language Processing}, pages 17889--17904, Miami, Florida, USA. Association for Computational Linguistics.

\bibitem[{Liu et~al.(2023)Liu, Yen, Huang, and Chen}]{contribution_sum}
Meng-Huan Liu, An-Zi Yen, Hen-Hsen Huang, and Hsin-Hsi Chen. 2023.
\newblock Contributionsum: Generating disentangled contributions for scientific papers.
\newblock In \emph{Proceedings of the 32nd ACM International Conference on Information and Knowledge Management}, pages 5351--5355.

\bibitem[{Liu et~al.(2019)Liu, Ott, Goyal, Du, Joshi, Chen, Levy, Lewis, Zettlemoyer, and Stoyanov}]{Liu2019RoBERTaAR}
Yinhan Liu, Myle Ott, Naman Goyal, Jingfei Du, Mandar Joshi, Danqi Chen, Omer Levy, Mike Lewis, Luke Zettlemoyer, and Veselin Stoyanov. 2019.
\newblock \href {https://api.semanticscholar.org/CorpusID:198953378} {Roberta: A robustly optimized bert pretraining approach}.
\newblock \emph{ArXiv}, abs/1907.11692.

\bibitem[{Luu et~al.(2021)Luu, Wu, Koncel-Kedziorski, Lo, Cachola, and Smith}]{explainingrelationships}
Kelvin Luu, Xinyi Wu, Rik Koncel-Kedziorski, Kyle Lo, Isabel Cachola, and Noah~A. Smith. 2021.
\newblock \href {https://arxiv.org/abs/2002.00317} {Explaining relationships between scientific documents}.
\newblock \emph{Preprint}, arXiv:2002.00317.

\bibitem[{Martin-Boyle et~al.(2024)Martin-Boyle, Tyagi, Hearst, and Kang}]{MartinBoyle2024ShallowSO}
Anna Martin-Boyle, Aahan Tyagi, Marti~A. Hearst, and Dongyeop Kang. 2024.
\newblock \href {https://api.semanticscholar.org/CorpusID:267751245} {Shallow synthesis of knowledge in gpt-generated texts: A case study in automatic related work composition}.
\newblock \emph{ArXiv}, abs/2402.12255.

\bibitem[{Meng et~al.(2020)Meng, Zhang, Huang, Xiong, Ji, Zhang, and Han}]{meng-etal-2020-text}
Yu~Meng, Yunyi Zhang, Jiaxin Huang, Chenyan Xiong, Heng Ji, Chao Zhang, and Jiawei Han. 2020.
\newblock \href {https://doi.org/10.18653/v1/2020.emnlp-main.724} {Text classification using label names only: A language model self-training approach}.
\newblock In \emph{Proceedings of the 2020 Conference on Empirical Methods in Natural Language Processing (EMNLP)}, pages 9006--9017, Online. Association for Computational Linguistics.

\bibitem[{Palani et~al.(2023)Palani, Naik, Downey, Zhang, Bragg, and Chang}]{palani2023relatedly}
Srishti Palani, Aakanksha Naik, Doug Downey, Amy~X. Zhang, Jonathan Bragg, and Joseph~Chee Chang. 2023.
\newblock \href {https://doi.org/10.1145/3544548.3580841} {Relatedly: Scaffolding literature reviews with existing related work sections}.
\newblock In \emph{Proceedings of the 2023 CHI Conference on Human Factors in Computing Systems}, CHI '23, New York, NY, USA. Association for Computing Machinery.

\bibitem[{Portenoy et~al.(2022)Portenoy, Radensky, West, Horvitz, Weld, and Hope}]{bridger}
Jason Portenoy, Marissa Radensky, Jevin West, Eric Horvitz, Daniel Weld, and Tom Hope. 2022.
\newblock \href {https://arxiv.org/abs/2108.05669} {Bursting scientific filter bubbles: Boosting innovation via novel author discovery}.
\newblock \emph{Preprint}, arXiv:2108.05669.

\bibitem[{Renduchintala et~al.(2024)Renduchintala, Bhatia, and Ramakrishnan}]{renduchintala2024smart}
HSVNS~Kowndinya Renduchintala, Sumit Bhatia, and Ganesh Ramakrishnan. 2024.
\newblock Smart: Submodular data mixture strategy for instruction tuning.
\newblock In \emph{Findings of the Association for Computational Linguistics ACL 2024}, pages 12916--12934.

\bibitem[{Sener and Savarese(2017)}]{sener2017active}
Ozan Sener and Silvio Savarese. 2017.
\newblock Active learning for convolutional neural networks: A core-set approach.
\newblock \emph{arXiv preprint arXiv:1708.00489}.

\bibitem[{Shi et~al.(2023)Shi, Gao, Zhang, Chen, Chen, Ren, and Ren}]{ur3wg}
Zhengliang Shi, Shen Gao, Zhen Zhang, Xiuying Chen, Zhumin Chen, Pengjie Ren, and Zhaochun Ren. 2023.
\newblock \href {https://doi.org/10.18653/v1/2023.findings-emnlp.385} {Towards a unified framework for reference retrieval and related work generation}.
\newblock In \emph{Findings of the Association for Computational Linguistics: EMNLP 2023}, pages 5785--5799, Singapore. Association for Computational Linguistics.

\bibitem[{Str\"{o}hle et~al.(2023)Str\"{o}hle, Campos, and Jatowt}]{Strhle2023}
Thomas Str\"{o}hle, Ricardo Campos, and Adam Jatowt. 2023.
\newblock \href {https://doi.org/10.1007/s41060-023-00434-4} {Contrastive text summarization: a survey}.
\newblock \emph{International Journal of Data Science and Analytics}, 18(4):353–367.

\bibitem[{Tindale(2020)}]{tindale2020self}
Christopher~W Tindale. 2020.
\newblock Self-deliberation and the strategy of the pseudo-dialogue.
\newblock \emph{Co-herencia}, 17(32):159--178.

\bibitem[{Tseng et~al.(2024)Tseng, Huang, Hsiao, Hsu, Foo, Huang, and Chen}]{tseng2024two}
Yu-Min Tseng, Yu-Chao Huang, Teng-Yun Hsiao, Yu-Ching Hsu, Jia-Yin Foo, Chao-Wei Huang, and Yun-Nung Chen. 2024.
\newblock Two tales of persona in llms: A survey of role-playing and personalization.
\newblock \emph{arXiv preprint arXiv:2406.01171}.

\bibitem[{Wang et~al.(2024{\natexlab{a}})Wang, Li, Li, Long, Tang, and Wang}]{dir}
Pancheng Wang, Shasha Li, Dong Li, Kehan Long, Jintao Tang, and Ting Wang. 2024{\natexlab{a}}.
\newblock Disentangling instructive information from ranked multiple candidates for multi-document scientific summarization.
\newblock In \emph{Proceedings of the 47th International ACM SIGIR Conference on Research and Development in Information Retrieval}, pages 2028--2037.

\bibitem[{Wang et~al.(2024{\natexlab{b}})Wang, Zhang, Robinson, Loeb, and Demszky}]{bridge}
Rose~E. Wang, Qingyang Zhang, Carly Robinson, Susanna Loeb, and Dorottya Demszky. 2024{\natexlab{b}}.
\newblock \href {https://arxiv.org/abs/2310.10648} {Bridging the novice-expert gap via models of decision-making: A case study on remediating math mistakes}.
\newblock \emph{Preprint}, arXiv:2310.10648.

\bibitem[{Wang et~al.(2024{\natexlab{c}})Wang, Mao, Wu, Ge, Wei, and Ji}]{wang-etal-2024-unleashing}
Zhenhailong Wang, Shaoguang Mao, Wenshan Wu, Tao Ge, Furu Wei, and Heng Ji. 2024{\natexlab{c}}.
\newblock \href {https://doi.org/10.18653/v1/2024.naacl-long.15} {Unleashing the emergent cognitive synergy in large language models: A task-solving agent through multi-persona self-collaboration}.
\newblock In \emph{Proceedings of the 2024 Conference of the North American Chapter of the Association for Computational Linguistics: Human Language Technologies (Volume 1: Long Papers)}, pages 257--279, Mexico City, Mexico. Association for Computational Linguistics.

\bibitem[{Wei et~al.(2022)Wei, Wang, Schuurmans, Bosma, Xia, Chi, Le, Zhou et~al.}]{wei2022chain}
Jason Wei, Xuezhi Wang, Dale Schuurmans, Maarten Bosma, Fei Xia, Ed~Chi, Quoc~V Le, Denny Zhou, et~al. 2022.
\newblock Chain-of-thought prompting elicits reasoning in large language models.
\newblock \emph{Advances in neural information processing systems}, 35:24824--24837.

\bibitem[{Xiao et~al.(2023)Xiao, Liu, Zhang, and Muennighoff}]{bge_embedding}
Shitao Xiao, Zheng Liu, Peitian Zhang, and Niklas Muennighoff. 2023.
\newblock \href {https://arxiv.org/abs/2309.07597} {C-pack: Packaged resources to advance general chinese embedding}.
\newblock \emph{Preprint}, arXiv:2309.07597.

\bibitem[{Yao et~al.(2023)Yao, Yu, Zhao, Shafran, Griffiths, Cao, and Narasimhan}]{yao2023tree}
Shunyu Yao, Dian Yu, Jeffrey Zhao, Izhak Shafran, Tom Griffiths, Yuan Cao, and Karthik Narasimhan. 2023.
\newblock Tree of thoughts: Deliberate problem solving with large language models.
\newblock \emph{Advances in neural information processing systems}, 36:11809--11822.

\bibitem[{Zhang et~al.(2024)Zhang, Lu, Yang, Zhou, Wan, and Zhang}]{Zhang2024FromCT}
Ming Zhang, Jiyu Lu, Jiahao Yang, Jun Zhou, Meilin Wan, and Xuejun Zhang. 2024.
\newblock \href {https://api.semanticscholar.org/CorpusID:268175720} {From coarse to fine: Enhancing multi-document summarization with multi-granularity relationship-based extractor}.
\newblock \emph{Inf. Process. Manag.}, 61:103696.

\end{thebibliography}

\appendix

\section{Experimental Settings}
\label{appendix: settings}
\par We choose \texttt{Llama-3.1-Nemotron-70B-Instruct-\\HF}, an \textit{open-source model}, as the base model for all experiments. We sample from the top 1\% of the tokens and set the temperature between $(0,0.5)$ to trade-off between deterministic and creative generation based on the nature of the given task (same setting across all samples):
\begin{itemize}
    \item \underline{Paper} generates arguments: $0.3$
    \item \underline{Paper} whether evidence is relevant: $0$
    \item \underline{Paper} presents its argument: $0.1$
    \item \underline{Paper} responds to the opposition's argument: $0.4$
    \item \underline{Paper} revises its argument: $0.4$
    \item \underline{Moderator} generating subtopics: $0.3$
    \item \underline{Moderator} determines whether to expand the debate note: $0.1$
    \item \underline{Moderator} summarizes a debate path: $0.4$
\end{itemize}

\par We set the number of retrieved segments $\delta = 5$ so that we can gather a sufficient amount of evidence while not overwhelming the debate with long-context. We set the number of generated subtopics $k = 3$, for covering a reasonable breadth of topics while minimizing redundancy. Finally, we set the maximum debate tree depth $l=3$ for adequate exploration. We use vLLM \cite{kwon2023efficient} for distributed and constrained generation on four \texttt{NVIDIA A100} GPUs.

\section{Baselines}
\label{appendix:baselines}
\par We compare Tree-of-Debate (ToD) with the following prompting-based baseline methods. 
We use the same base language model for all comparisons. 
\begin{itemize}[leftmargin=*]
    \item \textbf{Single-stage}: We prompt an LLM with the title, abstract and introduction sections of both focus and opposition papers. We prompt the model to directly generate a contrastive summary of the two papers \cite{MartinBoyle2024ShallowSO}. 
    \item \textbf{Two-stage}: We first instruct an LLM to individually summarize each paper based on the title, abstract and introductions. We then use the generated summaries to prompt the model to generate a contrastive summary \cite{Zhang2024FromCT}. 
\end{itemize}

To contextualize improvements from each component in Tree-of-Debate we construct the following ablative methods:
\begin{itemize}
    \item \textbf{ToD (No Tree)}: We remove the tree structure from Tree-of-Debate by merging child arguments into one. We do so by concatenating the topics and descriptions of the child subtopics and tag them to distinguish the topics. In each debate round, the model is prompted with the combined subtopic and its corresponding description. 
    \item \textbf{ToD (No SD)}: We remove self-deliberation (SD) to test the impact of iterative retrieval based on debate progression. We do so by prompting the model with title, abstract, and introduction of each paper instead of retrieving based on the subtopic. 
\end{itemize}

\section{Evaluation Metrics}
\label{appendix:eval_metrics}
\par The same domain-experts from Section \ref{sec:dataset} manually evaluate each sample in-depth, assessing various qualities of the summaries:

\begin{itemize}
    \item \textbf{Factuality}: \textit{How factual is the summary?} Each sentence is given a 1/0 binary score for factual or not, and the scores are averaged across the summary.
    \item \textbf{Completeness}: \textit{Is the summary comprehensive and complete?} This is evaluated using the following Likert scale:
    \begin{enumerate}
        \setcounter{enumi}{-1}
        \item Not at all, the summary misses (MULTIPLE) major points.
        \item No, the summary misses a (SINGULAR) major point.
        \item Somewhat, the summary misses minor points.
        \item Yes, the summary covers the major points, but still is not what I would expect.
        \item Very comprehensive, the summary covers the major points.
    \end{enumerate}
    \item \textbf{Contextualization}: 
    \textit{Does the summary explain and/or justify the posed similarities/differences between the papers, as opposed to just mentioning them?}
    \begin{enumerate}
        \setcounter{enumi}{-1}
        \item Not at all, the summary is simply extractive-- it just seems to take different subtopics from each paper and doesn’t synthesize them– no justification behind similarities and differences.
        \item No, the summary attempts at some level of synthesis, but it is not meaningful.
        \item Somewhat, the summary attempts at synthesizing at most one point.
        \item Yes, the summary contains meaningful synthesis but only for a minority of points.
        \item  Strongly contextualized, the summary contains meaningful synthesis across all major points.
    \end{enumerate}
\end{itemize}

\par We then perform min-max normalization to convert the Likert Scale-based scores to a value between 0 and 1 (e.g., $(x-0)/4$) across all annotators.

\section{Domain-Expert Profiles}
\label{appendix: annotators}
\par Given that novelty comparison between papers is a highly specialized and expensive task, we gather five domain experts to both collect and annotate our dataset, as well as evaluate Tree-of-Debate's generated summaries based on their respective samples. Each domain expert is a graduate student with $3+$ years of research experience in a specialized area:
\begin{enumerate}
    \item \textbf{Domain Expert \#1:} A third-year PhD in computer science with ten publications; research expertise is text mining and data mining.
    \item \textbf{Domain Expert \#2:} A third-year PhD in aerospace engineering with two publications; research expertise is in electric propulsion.
    \item \textbf{Domain Expert \#3:} A second-year PhD (with two years of a research-track Masters) in electrical engineering with four publications; research expertise is in in-memory computing and wireless communications.
    \item \textbf{Domain Expert \#4:} A first-year PhD (with two years of a research-track Masters) in computer science with six publications; research expertise is data-efficient natural language processing.
    \item \textbf{Domain Expert \#5:} A first-year PhD (with two years of a research-track Masters) in computer science with twenty-five publications; research expertise is large language model training and efficiency.

\section{Human Annotation \& Evaluation Agreement}
\label{appendix:agreeement}

\par{\textbf{Annotation Agreement.}} We note that in our dataset, the annotations for method/task and cites/not-cited are not subjective, and have clear guidelines when two papers differ by task or methodology, or have a citation link between them (Section \ref{sec:dataset}). Nonetheless, we had the NLP domain-expert relabel the data/text mining paper pairs (and vice versa), and found a 100\% agreement score between the original labels and re-annotated labels.

\par{\textbf{Evaluation Agreement.}} Due to the nature of the task, it is challenging to find multiple domain-experts that study the same fine-grained research topics (e.g., hall thrusters + electric propulsion in Aerospace Engineering; wireless sensing + in-memory computing for Electrical Engineering). However, the scale and degree of overlap between certain computer science subfields allows for multiple domain-experts to evaluate at least a subset of the same papers. Thus, the NLP and data mining domain-experts evaluated the summaries for a subset of each other's papers which they felt confident in their comprehension of. Based on these results, Breadth and Contextualization both have \textbf{80\% agreement rate} and \textbf{95\% ordinal agreement index}. We show factuality's intra-class correlation (ICC) scores in Table \ref{tab:icc_factuality}, which indicate that factuality has ``\textbf{good reliability}'' and \textbf{close to ``excellent reliability''}.

\begin{table}[h]
\centering
\begin{tabular}{lc}
\toprule
\textbf{ICC} & \textbf{Factuality} \\
\midrule
\textbf{ICC1k} & 0.855802 \\
\textbf{ICC2k} & 0.860777 \\
\textbf{ICC3k} & 0.891537 \\
\bottomrule
\end{tabular}
\caption{Intra-class correlation (ICC) scores for NLP/Data Mining factuality evaluation.}
\label{tab:icc_factuality}
\end{table}

\begin{table*}[h!]
\scriptsize
\centering
\begin{tabular}{p{2.5cm}p{6cm}p{6cm}}
\toprule
\textbf{Initial Debate Query: text classification} & \texttt{\textbf{BAAI/bge-large-en-v1.5}} & \texttt{\textbf{intfloat/e5-large-v2}} \\
\midrule
\textbf{Rank \#1 Segment} 
& Introduction Text classification is a fundamental task in Natural Language Processing (NLP) that enables automatic labeling of massive text corpora, which is necessary in many downstream applications [...] Prior works train text classifiers in a fully-supervised manner [...] that requires a substantial amount of training data, which is expensive and time-consuming, especially in emerging domains [...]
& , w|sj ]] as a list of words. Class Representations. Weakly supervised text classification problems can be initially approached by getting class representations using a pre-trained language model \\
\midrule
\textbf{Rank \#2 Segment} 
& However, existing approaches treat different levels of text granularity (documents, sentences, or words) independently, disregarding inter-granularity class disagreements and the context identifiable exclusively through joint extraction. In order to tackle these issues, we introduce MEGClass, an extremely weakly supervised text classification method that leverages Mutually-Enhancing Text Granularities. MEGClass utilizes coarse and fine-grained context signals obtained by jointly considering a document's most class-indicative words and sentences
& Introduction Text classification is a fundamental task in Natural Language Processing (NLP) that enables automatic labeling of massive text corpora, which is necessary in many downstream applications [...] Prior works train text classifiers in a fully-supervised manner [...] that requires a substantial amount of training data, which is expensive [...] \\
\midrule
\textbf{Rank \#3 Segment} 
& w|sj ]] as a list of words. Class Representations. Weakly supervised text classification problems can be initially approached by getting class representations using a pre-trained language model
& Finally, handling class hierarchies may improve MEGClass's fine-grained classification abilities. Ethics Statement [...] text classification is a standard problem across NLP applications and basing it on extremely weak supervision helps as a barrier to any user-inputted biases. [...] any hidden biases that exist within the pre-trained language models [...] \\
\midrule
\textbf{Rank \#4 Segment} 
& The pseudotraining dataset is used to fine-tune a text classifier. 3. Experiments on seven datasets demonstrate that MEGClass outperforms existing weakly and extremely weakly supervised methods, significantly in long-document datasets and competitively in shorter, coarse-grained datasets.
& It then utilizes the most confident documents for each class to update its understanding of the respective class, ultimately constructing a pseudo dataset to fine-tune a text classifier. Sentence Representation: Class-Oriented Each word in a sentence is assigned a weight based on the similarity between its closest class and the average of all class representations, finally to derive a class-oriented sentence representation \\
\bottomrule
\end{tabular}
\caption{Comparison of top retrieved segments between different base retrieval models (BAAI/bge-large-en-v1.5 and intfloat/e5-large-v2) for the query: \textit{text classification}.}
\label{tab:retrieved_segments_comparison}
\end{table*}

\begin{table*}[!h]
\scriptsize
\centering
\begin{tabular}{lp{10cm}c}
\toprule
\textbf{Model} & \textbf{Summary Takeaways} & \textbf{Raw Scores} \\
\midrule

\textbf{Llama-3.1-8B} & 
- Focuses on pre-trained language models as the sole similarity; should have mentioned the supervision levels and presence of iterative feedback. \newline
- Correctly identifies the differences in textual granularities used by the two papers and its significance towards fine-grained classification \newline
- Correctly identifies the differences in how the different methods compute the different contextualized document representations \newline
- Correctly identifies that LOTClass has a more robust self-training approach; also interestingly mentions that LOTClass uses larger-scale datasets which is a good point to mention (albeit worded superfluously) \newline
- Implicitly frames LOTClass more as a work for fine-grained text classification (e.g., ``LOTClass's self-training approach is more robust and efficient in handling fine-grained text classification tasks.'') 
& 2 / 4 / 0.7273 \\
\midrule

\textbf{Qwen2.5-32B} & 
- Correctly mentions the goal of weak supervision, but does not explicitly state that both use the same type of supervision (surface class labels); only mentions this for LOTClass \newline
- Correctly identifies the differences in textual granularities used by the two papers and its significance towards fine-grained classification (good that it mentions the performance on specific fine-grained datasets) \newline
- Mentions LOTClass’s strong performance on the datasets; not as important for a comparative analysis since these are non-overlapping with MEGClass $\rightarrow$ may be leaning on abstract more for this point \newline
- Correctly identifies that LOTClass’s approach is more simple and thus more adaptable and efficient. \newline
- Did not mention MEGClass’s iterative feedback approach–relevant to LOTClass’s self-training approach. 
& 3 / 4 / 1 \\
\midrule

\textbf{Llama-3.1 Nemotron-70B} & 
- Correctly identifies that both methods utilize weak supervision \newline
- Correctly identifies that both methods utilize iterative refinement techniques (and mentions the specific approaches as well) \newline
- Correctly identifies the differences in textual granularities used by the two papers and its significance towards long document classification \newline
- Correctly mentions LOTClass’s simplicity – Bonus! \newline
- Provides further analysis on MEGClass’s robustness due to considering multiple granularities, while mentioning that LOTClass may be influenced by PLM biases due to relying solely on class labels (single granularity) \newline
- Great analysis towards the end: ``the trade-offs between the two approaches, with MEGClass offering enhanced contextual understanding at the cost of increased complexity, and LOTClass providing simplicity and generalizability at the potential expense of nuance in classification.''
& 4 / 4 / 1 \\
\bottomrule
\end{tabular}
\caption{Summary takeaways and scoring across different base models for comparative text classification analysis. Score format is: breadth / context / fact.}
\label{tab:model_takeaway_scores}
\end{table*}

\begin{table*}[ht!]
\scriptsize
\centering
\begin{tabular}{lp{13cm}p{1cm}}
\toprule
\textbf{D} & \textbf{Summary Takeaways} & \textbf{Scores} \\
\midrule

\textbf{1} & 
- Correctly identifies that both methods focus on weak supervision, but claims that LOTClass is the only method that uses extremely weak supervision (both methods do) \newline
- Vague statement: ``This granularity-agnostic property makes LOTClass more versatile, particularly in scenarios where multi-granularity information is limited.'' \newline
- Somewhat surface-level points; the summary seems to mainly draw on the abstracts of each work. For example, ``MEGClass uniquely employs mutually-enhancing text granularities'' $\rightarrow$ a more contextualized summary would explain the different granularities (word, sentence, and doc-level) and which ones overlap with LOTClass (document-level). \newline
- Another high-level statement: ``Ultimately, the papers' distinct strengths and weaknesses underscore the complexity of text classification with minimal supervision, highlighting the importance of continued research in this area'' \newline
- Overall, the correct points are brought up (e.g., MEGClass’s iterative feedback vs LOTClass’s self-training), but they seem a bit surface level and do not indicate ``deep understanding'' of the works’ similarities and differences. 
& 4 / 2 / 0.7272 \\
\midrule

\textbf{2} & 
- Great that it is mentioned that both papers tackle weak supervision \textit{and} use PLMs \newline
- Misses that both have a similarity of having an iterative approach (but still brings it up within the dissimilarities since they are \textit{different} approaches to iterative feedback) \newline
- Correctly identifies the differences in textual granularities used by the two papers and its significance towards fine-grained classification \newline
- Correctly brought up simplicity: ``LOTClass prioritizing simplicity and MEGClass balancing efficiency with contextual understanding'' \newline
- Some factuality issues: ``LOTClass's self-training process effectively mitigates biases through adaptive label utilization [...] MEGClass contributes more significantly to the topic of Efficient Self-Training for Text Classification'' $\rightarrow$ not a fair argument to make 
& 3 / 4 / 0.7 \\
\midrule

\textbf{3} & 
- Correctly identifies that both methods utilize weak supervision \newline
- Correctly identifies that both methods utilize iterative refinement techniques (and mentions the specific approaches as well) \newline
- Correctly identifies the differences in textual granularities used by the two papers and its significance towards long document classification \newline
- Correctly mentions LOTClass’s simplicity – Bonus! \newline
- Provides further analysis on MEGClass’s robustness due to considering multiple granularities, while mentioning that LOTClass may be influenced by PLM biases due to relying solely on class labels (single granularity) \newline
- Great analysis towards the end: ``the trade-offs between the two approaches, with MEGClass offering enhanced contextual understanding at the cost of increased complexity, and LOTClass providing simplicity and generalizability at the potential expense of nuance in classification.'' 
& 4 / 4 / 1 \\
\bottomrule
\end{tabular}
\caption{Depth-wise (\textbf{D}) evaluation of summary takeaways (provided by domain-experts) and their corresponding scores. Score format is: breadth / context / factuality.}
\label{tab:depth_summary_scores}
\end{table*}

\section{Case Study on Base Retrieval Embedding Model}
\label{appendix:retrieval}

\par For the base retrieval embedding model, we used \texttt{BAAI/bge-large-en-v1.5}, which is one of the most popular retrieval embedding models that is specifically designed for retrieval-augmented LLMs and dense retrieval tasks. As a supplementary experiment, we show a qualitative comparison between BGE and e5 in Table \ref{tab:retrieved_segments_comparison}, where we include the top four retrieved segments given the initial topic query (using the MEGClass/LOTClass paper pair used in Table \ref{tab:summary}). This allows us to determine the difference in retrieval quality independent of any other influencing factors (e.g., LLM temperature).

\par In Table \ref{tab:retrieved_segments_comparison}, we can see that \textbf{\textit{two of the top-4 segments}} overlap between the two embedding models (BGE segments \#1 and \#3; e5 segments \#1 and \#2). Additionally, BGE’s segment \#2 and \#4 discuss the use of a pseudo-training dataset and class-indicative terms, \textbf{\textit{similar to e5’s segment \#4}}. However, we can see that e5’s segment \#3 primarily discusses the ethical concerns (e.g., bias in pre-trained language models). This is not as relevant to the core query topic of ``text classification'', hence making \textbf{the retrieved segments of BGE slightly more preferable}.

\section{Case Study on Base LLM}
\label{appendix:base_llm}

\par We show a qualitative case study for the same paper pair used in Table \ref{tab:summary}, but across base model type and size. Given the response character limit, we include the specific feedback from the annotator on the generated summaries and their corresponding scores instead in Table \ref{tab:model_takeaway_scores}.

\par Overall, we can see that the \textbf{contextualization} performance across different base models and model sizes \textbf{is quite consistent}, with \textbf{smaller model sizes primarily impacting the breadth and factuality of the summary}. It seems that using the \textbf{larger model} (70B) results in Tree-of-Debate exhibiting \textbf{better breadth performance} (can identify all the major comparison points) and \textit{drawing out more meaningful analysis}.

\section{Influence of Debate Depth}
\label{appendix: depth}

\par \par Table \ref{tab:depth_summary_scores} presents a qualitative case study on how the summaries evolve as we vary the depth from 1 to 3. \textbf{As depth increases, both factuality and contextualization generally increase.} Based on the expert’s feedback, it seems that depth=1 produces shallow summaries that lean more on their surface-level knowledge of the paper’s abstract and high-level points--- leading to less contextualized summaries. This also impacts the factuality, leading to more superfluous/overly positive statements. However, this significantly improves as depth increases, where the \textit{personas are able to dive deeper into fine-grained points} and are able to further \textit{refine/tone-down their exaggerated arguments}.

\section{Error Analysis}
\label{appendix:error_analysis}
\par ToD’s \textbf{\textit{factuality}} mistakes are primarily due to confusing facts between different papers. For example, as ToD compares TreeInstruct \cite{kargupta2024instructassistllmbasedmultiturn} and BRIDGE \cite{bridge}, it mentions that both have ``distinct approaches to addressing interdependent mistakes''. The reality is that TreeInstruct explicitly addresses interdependent mistakes, while BRIDGE does not. According to the human annotator, this sentence received a factuality score of 0.

\par A \textbf{\textit{breadth}} mistake is when certain similarities or differences are excluded from the summary. For example, when comparing SMART \cite{renduchintala2024smart} and ``Active Learning for Convolutional Neural Networks'' \cite{sener2017active}, ToD misses a similarity between these papers that is relevant to the topic of ``efficient data selection''. Both papers represent data similarity as proximity in the embedding space – despite being an important detail in data-efficient research, this similarity is not included in the summary. For this reason, the human annotator gave the summary a 2/4 score for breadth (a breadth score of 2 indicates that the summary misses minor points).

\par ToD tends to make fewer \textbf{contextualization} mistakes, but the mistakes occur when differences are not explained properly. For example, when comparing Tree-of-Thoughts \cite{yao2023tree} (ToT) and Chain-of-Thought \cite{wei2022chain}, ToD recognizes that ToT uses      ``deliberate decision-making'' and ``a more structured decision-making process'', but never elaborates on the specifics of the decision-making process. Because the ``decision-making process'' was mentioned twice by ToD, it is important to explain the process clearly and succinctly. Hence, this is a contextualization mistake. According to the domain-expert, this summary received a 3/4 because the differences and similarities were not adequately explained.

\section{Comparative Analysis of Output Summaries: TreeInstruct vs. BRIDGE}
\label{appendix:summary}
\begin{table*}[]
    \footnotesize
    \centering
    \caption{Summaries from two methods: ToD (No Tree) -- the \textsc{Tree-of-Debate} ablation -- and the Two-Stage baseline. The papers being compared are TreeInstruct and BRIDGE: they develop LLM-based instructors that guide students towards the answers to their questions rather than give away the answers. The rest of the table follows in \ref{table: qualitative_study_appendix_part2}.}
    \begin{tabular}{|p{7cm}|p{7cm}|}
        \toprule
        \textbf{Two-Stage} & \textbf{ToD (No Tree)} \\
        \midrule
        Both papers focus on enhancing educational experiences through innovative technological approaches. Both utilize novel methodologies to address specific learning challenges (code debugging for Paper 0 and math remediation for Paper 1). Both papers emphasize the importance of real-world validation and case studies to assess the effectiveness of their proposed solutions. Topic Focus: Paper 0 focuses on code debugging with TreeInstruct, while Paper 1 addresses math mistake remediation with the Bridge method. \textcolor{red}{Methodological Approach: Paper 0 employs state space estimation and dynamic tree-based questioning, whereas Paper 1 utilizes cognitive task analysis and decision-making models.} Evaluation Metrics: Paper 0 measures performance on debugging datasets and student engagement, whereas Paper 1 assesses the preference for LLM responses with expert decisions and conducts lexical analysis. Target Audience: Paper 0's work is geared towards students of varying skill levels in code debugging, whereas Paper 1 focuses on 1st-5th grade students in Title I schools for math tutoring. While both papers contribute to the educational technology landscape, Paper 0's TreeInstruct offers a novel approach to code debugging through adaptive Socratic questioning, and Paper 1's Bridge method innovatively embeds expert thought processes into LLMs for math remediation. \textcolor{red}{Despite overlapping in their pursuit of enhancing educational experiences, the papers diverge significantly in their topics, methodologies, and target audiences, indicating a broad and diverse landscape in educational technology research.} & Papers on adaptive guidance systems debate similarities and differences for effective Socratic questioning. While both papers aim to enhance adaptive guidance, they differ in approach: \textcolor{red}{TreeInstruct (Paper 0) emphasizes adaptive, expert-agnostic planning, leveraging dynamic question trees and state space-based planning to address interdependent errors, whereas Bridge (Paper 1) focuses on expert-informed decision-making, incorporating domain-specific expert knowledge for more effective Socratic questioning.} A key difference lies in their handling of unforeseen errors, with TreeInstruct relying on dynamic adaptation and \textcolor{red}{Bridge on continuous expert annotation and model fine-tuning. }Another distinction is their applicability: \textcolor{red}{TreeInstruct is designed for diverse educational domains}, while Bridge initially focused on math education, later expanding to other domains. Both authors eventually converge on a hybrid approach, combining expert guidance with adaptive planning for a more comprehensive solution. Despite this, their initial differences in methodology and error handling strategies remain a focal point of the debate. The debate highlights the importance of both adaptive planning and expert guidance in effective adaptive guidance systems. Ultimately, the synthesis of both approaches offers a more robust solution for personalizing student learning experiences. \\
\bottomrule
    \end{tabular}
    \label{table: qualitative_study_appendix_part1}
\end{table*}

\begin{table*}[]
    \small
    \centering
    \caption{Summaries from two methods: ToD (No SD) -- the \textsc{Tree-of-Debate} ablation -- and \textsc{Tree-of-Debate}. The papers being compared are TreeInstruct and BRIDGE: they develop LLM-based instructors that guide students towards the answers to their questions rather than give away the answers. The rest of the table follows in \ref{table: qualitative_study_appendix_part1}.}
    \begin{tabular}{|p{7cm}|p{7cm}|}
        \toprule
        \textbf{ToD (No SD)} & \textbf{ToD (Final)} \\
        \midrule
        The papers share similarities in their focus on adaptive, student-centered approaches for mistake correction, leveraging AI-driven tools to enhance personalized learning experiences. Both papers emphasize the importance of context-aware feedback and expert guidance in promoting deeper understanding. However, the papers diverge in their novelties. Paper 0's paper, 'Instruct, Not Assist,' uniquely integrates adaptive, multi-turn Socratic questioning with state space estimation, \textcolor{red}{providing a more holistic and adaptable learning experience}. In contrast, Paper 1's paper, 'Bridging the Novice-Expert Gap,' focuses on modular, expert-guided decision-making, offering a more explicit and scalable approach to addressing complex, interdependent mistakes. \textcolor{ForestGreen}{While Paper 0's approach excels in handling ambiguous student responses, Paper 1's method provides more nuanced, human-centric guidance.} The papers also differ in their evaluation metrics, with Paper 0 emphasizing robust state space estimation and Paper 1 highlighting the importance of contextual relevance. Ultimately, \textcolor{red}{the papers present complementary approaches, with Paper 0's work prioritizing adaptability and Paper 1's work emphasizing expert-driven contextualization}. & The papers by Paper 0 and Paper 1 share similarities in their focus on adaptive error resolution strategies, emphasizing the importance of context-sensitive decision-making and expert knowledge in enhancing student learning outcomes. \textcolor{ForestGreen}{Both papers also highlight the limitations of current LLMs in educational settings} and propose novel methodologies to address these challenges. However, the papers diverge in their approaches to achieving these goals. \textcolor{ForestGreen}{Paper 0's paper, 'Instruct, Not Assist,' uniquely integrates dynamic planning and hierarchical questioning to provide a more nuanced understanding of student thought processes}, scaling to accommodate diverse student populations. In contrast, \textcolor{ForestGreen}{Paper 1's paper, 'Bridging the Novice-Expert Gap,' leverages expert decision-making models to inform adaptive error resolution strategies, offering a more domain-agnostic framework}. The papers also differ in their evaluation frameworks, with Paper 0's approach utilizing a state space-based planning algorithm and Paper 1's method incorporating transfer learning and domain-specific expert annotations. Furthermore, \textcolor{ForestGreen}{Paper 0 emphasizes the importance of adaptive Socratic questioning, while Paper 1 highlights the value of expert-guided decision-making in enhancing educational support.} Ultimately, the papers present distinct novelties in addressing the novice-expert gap, with \textcolor{ForestGreen}{Paper 0 focusing on adaptive structured planning and Paper 1 on context-aware expert decision embedding}. \\
\bottomrule
    \end{tabular}
    \label{table: qualitative_study_appendix_part2}
\end{table*}

Tables \ref{table: qualitative_study_appendix_part1} and \ref{table: qualitative_study_appendix_part2} contain comparative summaries from the baseline, ablations, and our final method on the papers, TreeInstruct \cite{kargupta2024instructassistllmbasedmultiturn} and BRIDGE \cite{bridge}--- Papers 0 and 1, respectively. Below, we qualitatively compare each summary, pointing out the weaknesses and strengths, and show how our method is able to address all the issues brought up in the baseline summaries.

The top left contains the Two-Stage baseline. The Two-Stage baseline tends to contain near-copy phrases from the paper, resulting in \textbf{an overly specific, extractive and unnatural summary} (an example is the first line highlighted in red: ``Methodological Approach: Paper 0 employs...''). As a result, the differences that are extracted are not explained very well, requiring more work to understand the terminology-heavy summary. It also makes vague claims near the end of summaries (example is the second line highlighted in red: ``Despite overlapping in their pursuit of enhancing educational experiences...''). \textbf{The overall structure results in a suboptimal summary}.

Next, the top right box contains the summary for ToD (No Tree). The use of the debate format improves the quality of the generated claims. Unlike in the Two-Stage summary, it does not contain many extractive phrases, however \textbf{the structure of the debate is still fine-grained to coarse-grained}. Intuitively, the summaries should develop coarse-grained claims into fine-grained arguments. Moreover, there are \textbf{slight hallucinations} (examples are in the second and third lines highlighted in red: ``Bridge on continuous expert...'' and ``TreeInstruct is designed...''). Still, the conclusion (last sentence) of the summary is not as vague as the conclusion from Two-Stage, but it still does not capture the intricacies of the two methods well enough.

Subsequently, the summary for ToD (No SD) is on the bottom left. The benefits of the tree are drastic, as the summary starts by discussing the high-level summaries, and breaks down the individual fine-grained differences. This is much less extractive and more abstractive. \textbf{Using the tree structure along with the debate allows each argument to be explored further}-- this is evident as after each claim, an explanation of why it matters follows (example is the line highlighted in green: ``While Paper 0's approach excels in...''). Still, a few of these explanations are vague and \textbf{do not reveal the true underlying motivation of the claims} (highlighted in red).

Finally, the summary for ToD (our final method) is in the bottom right box. With the self-deliberation, it was able to extract a short phrase of the motivation behind both works (the ``limitations of current LLMs in educational settings''). The arguments are developed from \textbf{high-level claims to low-level}, technical concepts. The \textbf{facts are correctly identified} and do not contain any hallucinations. Moreover, the explanations preceding the claims also \textbf{reveal the underlying motivation} behind the specific novelty. Finally, the concluding sentence explains the exact difference between the two works.

\section{Tree Example}
\label{sec:tree_example}

In Table \ref{appendix:example_case} below, we provide an example of a debate path generated by \textit{Tree-of-Debate}. We would utilize this type of path in-context for the final summarization step. Given input topic: \textbf{Helping Students Fix their Mistakes}.

\begin{table*}
\caption{Example of Debate Tree Path for \cite{kargupta2024instructassistllmbasedmultiturn} and \cite{bridge}.}
\label{appendix:example_case}
\begin{tabular}{|p{15cm}|}
\hline
\textbf{(a) Level 1 Child Argument: Personalized Error Resolution Strategies.} \newline
Debate the effectiveness of adaptive, student-centered approaches in resolving errors, focusing on the role of dynamic question trees and expert decision-making models. This subtopic encourages discussion on the importance of tailoring guidance to individual students’ needs and knowledge gaps.

\begin{itemize}[leftmargin=1em]
    \item \textbf{Author 0’s Argument: Adaptive Guidance Enhances Personalization via Nuanced Student Modeling.} \newline
    By integrating TreeInstruct’s dynamic question tree with a nuanced student modeling framework, our approach provides more effective personalized error resolution strategies than Bridge. This integration enables TreeInstruct to capture the complexities of student thought processes, addressing both independent and dependent mistakes concurrently, while also scaling to accommodate large, diverse student populations. In contrast, Bridge’s reliance on predetermined error types and remediation strategies may limit its versatility in complex, multi-bug scenarios.

    \item \textbf{Author 1’s Argument: Expert-Guided Decision-Making for Adaptive, Context-Sensitive Error Resolution.} \newline
    Our approach leverages expert decision-making to inform adaptive, context-sensitive error resolution strategies, demonstrated through the extension of our Bridge model to accommodate complex, multi-bug scenarios. By incorporating nuanced expert thought processes, our method provides more effective personalized guidance than TreeInstruct’s adaptive question tree approach. This integration of expert guidance and adaptability addresses the opposition’s concerns regarding versatility and scalability in handling diverse student populations.
\end{itemize}

\vspace{1em}

\textbf{(a.i) Level 2 Child Argument: Adaptive Guidance in Error Resolution.} \newline
Debate the effectiveness of adaptive guidance in error resolution strategies, focusing on how each approach tailors feedback to individual students’ knowledge states. Discuss the benefits and limitations of each method.

\begin{itemize}[leftmargin=1em]
    \item \textbf{Author 0’s Argument: Adaptive Guidance with Latent Expertise for Personalized Error Resolution.} \newline
    By integrating the strengths of both approaches, our revised argument proposes a hybrid model that combines the adaptive guidance of TreeInstruct with the latent thought processes of expert decision-making. This fusion enables a more nuanced understanding of individual student needs, providing targeted support while maintaining scalability and adaptability. Ultimately, this hybrid approach offers a more effective and personalized error resolution strategy, surpassing the limitations of both adaptive guidance and expert decision-making alone.

    \item \textbf{Author 1’s Argument: Expert-Infused Adaptive Guidance for Error Resolution.} \newline
    Our revised approach integrates expert decision-making processes into adaptive guidance frameworks, enabling a more nuanced understanding of individual student needs and providing targeted support while maintaining scalability and adaptability. This hybrid model combines the strengths of both approaches, offering a more effective and personalized error resolution strategy. By incorporating expert-informed decision-making into adaptive guidance, we bridge the novice-expert knowledge gap more effectively than either approach alone.
\end{itemize}
\\
\hline
\end{tabular}
\end{table*}

\begin{table*}[h!]
    \scriptsize
    \centering
    \renewcommand{\arraystretch}{1.3}
    \caption{Snapshot of 5 rows of the dataset. The table is transposed in order to display it on the full page. Notice the diversity in topics, not just in machine learning, but also in other disciplines.}
    % \resizebox{\linewidth}{!}{
    \begin{tabular}{|p{1cm}|p{2.25cm}|p{2.25cm}|p{2.25cm}|p{2.25cm}|p{2.25cm}|}
    \hline
    Index & 1 & 2 & 3 & 4 & 5 \\
    \hline
    \textbf{Topic} & helping students understand their mistakes and misunderstandings & Continual pretraining of Bert for retrieval tasks & contrastive learning on graphs & hall thruster erosion & massive MIMO baseband processing \\ \hline
    \textbf{Paper \#1 arXiv} & https://arxiv.org/ pdf/2406.11709 & https://arxiv.org/ pdf/2004.12832 & https://arxiv.org/ pdf/2103.00113 & https://deepblue.lib. umich.edu/bitstream/ handle/2027.42... & https://arxiv.org/ pdf/2407.06755 \\ \hline
    \textbf{Paper \#1 Title} & Instruct, Not Assist: LLM-based Multi-Turn Planning and Hierarchical Questioning for Socratic Code Debugging & ColBERT: Efficient and Effective Passage Search via Contextualized Late Interaction over BERT & Anomaly Detection on Attributed Networks via Contrastive Self-Supervised Learning & An Investigation of Factors Involved in Hall Thruster Wall Erosion Modeling & A 46 Gbps 12 pJ/b Sparsity-Adaptive Beamspace Equalizer for mmWave Massive MIMO in 22FDX \\ \hline
    \textbf{Paper \#1 Abstract} & Socratic questioning is an effective teaching... & Recent process in Natural Language Understanding... & Anomaly detection on attributed networks attracts & A hydrodynamic description of the plasma flow...  & We present a GlobalFoundries 22FDX FD-SOI... \\ \hline
    \textbf{Paper \#1 Introduction} & With the rapidly expanding conversational... & Over the past few years, the Information... & Attributed networks (a.k.a. attributed graphs... & As lifetime requirements desired for Hall thruster... & Fifth generation (5G) and beyond-5G wireless... \\ \hline
    \textbf{Paper \#2 arXiv} & https://arxiv.org/ pdf/2310.10648 & https://arxiv.org/ pdf/1810.04805 & https://arxiv.org/ pdf/2310.14525 & https://electricrocket .org/IEPC/IEPC-2007-151.pdf & https://arxiv.org/ pdf/1910.00756 \\ \hline
    \textbf{Paper \#2 Title} & Bridging the Novice-Expert Gap via Models of Decision-Making: A Case Study on Remediating Math Mistakes & BERT: Pre-training of Deep Bidirectional Transformers for Language Understanding & Graph Ranking Contrastive Learning: A Extremely Simple yet Efficient Method & Investigation of the SPT operation and discharge chamber wall erosion rate under increased discharge voltages & Beamspace Channel Estimation for Massive MIMO mmWave Systems: Algorithm and VLSI Design \\ \hline
    \textbf{Paper \#2 Abstract} & Scaling high-quality tutoring remain a major challenge... & We introduce a new language representation... & Graph contrastive Learning (GCL) has emerged as... &  New results of the thruster operation specifics & Millimeter-wave (mmWave) communication \\ \hline
    \textbf{Paper \#2 Introduction} & Human tutoring plays a critical role in accelerating student... & Language model pre-training has been shown to be effective... & Graph Neural Networks (GNNs) have become the standard & Nowadays it is reasonable to develop SPT with increased  & Millimeter-wave (mmWave) communication [2], [3]  \\ \hline
    \textbf{Method /Task} & 1 & 1 & 1 & 0 & 1 \\ \hline
    \textbf{Cite/No} & 1 & 1 & 0 & 0 & 0 \\ \hline
    
    \end{tabular}
    % }
    \label{tab: dataset_snapshot_in_appendix}
\end{table*}

\section{Dataset Specifications}
\label{appendix:dataset}
As mentioned earlier, \textsc{Tree-of-Debate}'s dataset contains 100 samples, and Table \ref{tab:dataset} specifies the breakdown. Each sample contains the following:
\begin{enumerate}
    \item Topic: a short, vague description of the theme of the two papers
    \item Paper \#1 arXiv Link
    \item Paper \#1 Title
    \item Paper \#1 Abstract
    \item Paper \#1 Introduction
    \item Paper \#2 arXiv Link
    \item Paper \#2 Title
    \item Paper \#2 Abstract
    \item Paper \#2 Introduction
    \item Method/Task: 0 if the papers differ in methodology (but have the same task) and 1\newline if the papers differ in the task (but the methodology is generally the same)
    \item Cite/No: 0 if the papers do not cite each other, and 1 if the papers cite each other.
\end{enumerate}

We provide the dataset (as a tab-separated file) in our code repository: \href{https://github.com/pkargupta/tree-of-debate/tree/main/dataset}{https://github.com/pkargupta/tree-of-debate/tree/main/dataset}. Table \ref{tab: dataset_snapshot_in_appendix} contains a subset of rows from the dataset.

\par \textbf{The pairs were identified manually by each annotator based on their own domain expertise and which papers would result in a ``meaningful comparative analysis''}---in other words, which pairs of papers that they think would or should naturally be compared (e.g., papers with similar goals, a new work that expands upon an old work, etc). This allows us to ensure that the papers in the same ``Cited'' or ``Not Cited'' pairs---especially for ``Not Cited''--- are truly relevant, and the annotator already has an understanding of the gold-standard comparative analysis. Hence, even the combinations of papers themselves are influenced by domain-specific knowledge--- not simply chosen by randomized pair sampling.

\end{enumerate}
\section{Prompts}
\label{appendix: prompts}
We provide each of the prompts used in our framework in the following tables.

\begin{table*}[h]
\centering
\begin{tabularx}{\textwidth}{|X|}
\hline
You are a fair and balanced moderator of a debate between two authors determining their respective novel contributions towards the following topic:

Topic: \texttt{<topic>} \\
Topic Description: \texttt{<topic description>}
\\\\
Here are the two papers and their claimed novel contributions with corresponding evidence:
\\\\
Author 0 Paper Title: \texttt{<Author 0 Paper Title>}\\
Author 0 Paper Abstract: \texttt{<Author 0 Paper Abstract>}\\
Author 0 Paper's Contribution \#1: \texttt{<contribution statement>}: \texttt{<contribution topic>}\\
Author 0 Paper's Contribution \#1 Evidence: \texttt{<contribution evidence>}\\
... (more evidence and contributions)

Author 1 Paper Title: \texttt{<Author 0 Paper Title>}\\
Author 1 Paper Abstract: \texttt{<Author 0 Paper Abstract>}\\
Author 1 Paper's Contribution \#1: \texttt{<contribution statement>}: \texttt{<contribution topic>}\\
Author 1 Paper's Contribution \#1 Evidence: \texttt{<contribution evidence>}\\
... (more evidence and contributions)
\\\\
Based on each of the author's claimed novelties, evidence, and counter-evidence to each other's arguments, you must determine the most meaningful, diverse set of subtopics within the parent topic, "Topic", which best cover the types of contributions each of the papers make. Remember that for each of your selected topics, the papers will be debating which of them makes the better contribution towards the topic. Hence, for each of your subtopics, cite the integer IDs of any relevant contributions from Author 0 or Author 1. At least one of these lists should be non-empty. Overall, our goal is to identify how novel Author 0's paper's contributions towards topic "Topic" are by individually considering their contributions towards your subtopics. 
\\\\
Output your list subtopics (up to $k$) in the following format:
    "topic\_title": <should be a brief, 10-15 word string where the value is the title of your subtopic>, \\
    "topic\_description": <1-2 sentence string explaining the subtopic and what you feel would be most helpful for the papers to debate within the subtopic>, \\
    "author\_0\_relevant\_contributions": <list of integer IDs citing which contribution(s) from Author 0 would be most relevant to this subtopic; can be empty>, \\
    "author\_1\_relevant\_contributions": <list of integer IDs citing which contribution(s) from Author 1 would be most relevant to this subtopic; can be empty>\\
 \\ \hline
\end{tabularx}
\caption{Moderator prompt to generate new topics.}
\label{prompt_mod: generate_topics}
\end{table*}

\begin{table*}[h]
\centering
\begin{tabularx}{\textwidth}{|X|}
\hline
You are a moderator facilitating a debate in which two paper are debating who makes the better contribution towards the following topic:\\
Topic: \texttt{<topic>} \\
Topic Description: \texttt{<topic description>}
\\\\
\texttt{<conversation history between Author 0 and Author 1>}
\\\\
Below, you are given the previous set of arguments and the current set of arguments. 

previous arguments: \texttt{<set of arguments before debate>}

current arguments: \texttt{<set of arguments after debate>}
\\\\
You must determine whether progress is being made. DO NOT focus on the language being used. Focus on the content of the arguments. Specifically, determine the following (True or False for each): \\
1. progression\_of\_arguments: Are these arguments sufficiently different enough to necessitate further debate? Are there new, deeper concepts being discussed between the two sets of arguments? \\
2. meaningful\_questions: Within the debate history, each author acknowledges each other's arguments and may ask clarifying questions accordingly. Do you believe that the clarifying questions have not been sufficiently addressed already and would be important to answer through further debate? If there are no questions raised in the debate history by either author, return False. \\
3. clear\_winner: Do you believe that it is clear that one author has won the debate, and it does not need to be further deconstructured (in order to determine which components within each author's contributions are truly better)?
\\\\
Output your argument in the following format: \\
    "explanation": <2-5 sentence string to explain your reasoning about whether further debate is necessary when comparing the previous arguments and the current arguments>,\\
    "progression\_of\_arguments": <output a boolean; pick only one of "True" or "False" depending on the history, arguments, and your explanation above>, \\
    "meaningful\_questions": <output a boolean; pick only one of "True" or "False" depending on the history, arguments, and your explanation above>, \\
    "clear\_winner": <output a boolean; pick only one of "True" or "False" depending on the history, arguments, and your explanation above>
 \\ \hline
\end{tabularx}
\caption{Moderator prompt on whether to expand a debate node.}
\label{prompt_mod: is_expand}
\end{table*}

\begin{table*}[h]
\centering
\begin{tabularx}{\textwidth}{|X|}
\hline
Two authors are debating their respective novelties with respect to the following topic: \\
Topic: \texttt{<Topic>} \\
Author 0's paper title is: \texttt{<Author 0 paper title>} \\
Author 1's paper title is: \texttt{<Author 1 paper title>} \\

Here is a breakdown of their debates in tree format. At each tree node, we provide the "topic\_title" : "topic description", Author 0's corresponding argument and Author 1's corresponding argument:
\\\\
\texttt{<tree (example in Appendix \ref{sec:tree_example})>}
\\\\
Based on the debate breakdown, output a paragraph-long synthesis of the debate which summarizes the similarities and differences between the papers. Structure your summary with initially their similarities (which ideas/aspects overlap between the two papers?) to their differences (what makes the papers unique) in novelties. Focus more on the differences than the similarities.

 \\ \hline
\end{tabularx}
\caption{Moderator prompt to summarize the debate into a paragraph.}
\label{prompt_mod: summarize_debate}
\end{table*}

\begin{table*}[h]
\centering
\begin{tabularx}{\textwidth}{|X|}
\hline
You are the author of the paper, '{self.paper.title}'. The abstract of your work is: \texttt{<Paper Abstract>}.

You are debating another author on the novel contributions your work makes towards the following topic: \texttt{<Topic>}.

Below is a list of relevant evidence retrieved from your paper:\texttt{<Evidence in the form of sentence excerpts from papers>}. Based on the evidence, output a list of 1 to \texttt{<k>} DIVERSE, specific arguments for your position that are all supported by the evidence. Each argument should have a corresponding "argument\_title", which is a brief statement of your argument (e.g., Better Efficiency for Training), a "description" explaining your argument and mentioning specific excerpts from your evidence pool, and finally, a list of all "evidence" IDs, which are the integers of the evidence in the input list, that best support your argument. For example, if Evidence \#1 and \#2 best support your argument, then evidence should be [1,2] (depending on your argument, this list can have more or less than two items). Each argument should make a unique point.
\\\\
Output your list of arguments in the following format: \\
                "argument\_title": <should be a brief, 10-15 word string where the value is the argument\_title>, \\
                "description": <1-2 sentence string explaining the argument, including specific excerpts from the evidence pool>, \\
                "evidence": <list of integer IDs citing which evidence from the input list best support your argument>
 \\ \hline
\end{tabularx}
\caption{Persona prompt to generate arguments during the debate.}
\label{prompt_persona: generate_arguments}
\end{table*}

\begin{table*}[h]
\centering
\begin{tabularx}{\textwidth}{|X|}
\hline
Your objective is to check if a given evidence is relevant to a claim or not (relevancy means evidence that helps either support, refute, or clarify the given claim).

Claim: \texttt{Argument}\\
Description of Claim: \texttt{Argument Description}\\
Evidence: \texttt{Evidence supporting the argument}.\\
\\
Fill out the following schema:\\
"supports\_claim": <"Yes"/"No" if the evidence supports the claim>,\\
"refutes\_claim": <"Yes"/"No" if the evidence refutes the opposition's claim>\\
"clarifies\_claim": <"Yes"/"No" if the evidence clarifies the claim>,\\
"irrelevant\_to\_claim": <"Yes"/"No" if the evidence is irrelevant to the claim>,
 \\ \hline
\end{tabularx}
\caption{Persona prompt to determine relevant and irrelevant evidences.}
\label{prompt_persona: is_irrelevant_evidences}
\end{table*}

\begin{table*}[h]
\centering
\begin{tabularx}{\textwidth}{|X|}
\hline
You are the author of the paper, '\texttt{<Paper Title>}'. The abstract of your work is: \texttt{<Paper Abstract>}.
\\\\
You are debating another author (Opposition), whose work is titled, '\texttt{<Opposition Paper Title>}', and abstract is: \texttt{Opposition Paper Abstract}".
\
You are debating the other author on how and why your paper makes a better contribution towards the following topic: \\
Topic: \texttt{<topic>} \\
Topic Description: \texttt{<topic description>}
\\\\
Here are your claimed contributions towards the topic: \\
Author 0 Paper's Contributions \#1: \texttt{<argument>}: \texttt{<argument description>} \\
Author 0 Paper's Contribution Evidence \#1: \texttt{<evidence towards argument>} \\
Author 1's relevant evidence to potentially counter the quality of this contribution: \texttt{<counter evidence>} \\
... (more contributions and counter-evidence)
\\\\
Given the above, make an argument for a specific reason why your contributions towards the topic, Topic: \texttt{<topic>}, are better than the opposition's. If you feel that you do not contribute to the given topic or your contributions ARE NOT better than the opposition's, then state so by conceding to the opposition (e.g., 'I do not believe my paper makes a better contribution than yours') and explain why. 
 \\ \hline
\end{tabularx}
\caption{Persona prompt to \textit{present} its arguments.}
\label{prompt_persona: present_arguments}
\end{table*}

\begin{table*}[h]
\centering
\begin{tabularx}{\textwidth}{|X|}
\hline
You are the author of the paper, '\texttt{<Paper Title>}'. The abstract of your work is: \texttt{<Paper Abstract>}.
\\\\
You are debating another author (Opposition), whose work is titled, '\texttt{<Opposition Paper Title>}', and abstract is: \texttt{Opposition Paper Abstract}".
\
You are debating the other author on how and why your paper makes a better contribution towards the following topic: \\
Topic: \texttt{<topic>} \\
Topic Description: \texttt{<topic description>}
\\\\
Here are your claimed contributions towards the topic: \\
Author 0 Paper's Contributions \#1: \texttt{<argument>}: \texttt{<argument description>} \\
Author 0 Paper's Contribution Evidence \#1: \texttt{<evidence towards argument>} \\
Author 1's relevant evidence to potentially counter the quality of this contribution: \texttt{<counter evidence>} \\
... (more contributions and counter-evidence)
\\\\
Here is your conversation debate history with the opposition paper. You must respond to the last argument presented by your opposition in debate. A response may consist of (1) an acknowledgment of the opposition's previous response, (2) answering any of the questions about your paper brought up by the opposition, (3) asking any clarifying questions based on the opposition's claims and reasoning, (4) any clarifications of your own presented arguments based on the opposition, and/or (5) if you feel that the opposition's claim is strong and you do not have sufficient grounds to refute it, then a concession to your opposition.
\\\\
conversation\_history: \texttt{<conversation history>}
 \\ \hline
\end{tabularx}
\caption{Persona prompt to \textit{respond} to an argument.}
\label{prompt_persona: respond_to_argument}
\end{table*}

\begin{table*}[h]
\centering
\begin{tabularx}{\textwidth}{|X|}
\hline
You are the author of the paper, '\texttt{<Paper Title>}'. The abstract of your work is: \texttt{<Paper Abstract>}.
\\\\
You are debating another author (Opposition), whose work is titled, '\texttt{<Opposition Paper Title>}', and abstract is: \texttt{Opposition Paper Abstract}".
\
You are debating the other author on how and why your paper makes a better contribution towards the following topic: \\
Topic: \texttt{<topic>} \\
Topic Description: \texttt{<topic description>}
\\\\
Here are your claimed contributions towards the topic: \\
Author 0 Paper's Contributions \#1: \texttt{<argument>}: \texttt{<argument description>} \\
Author 0 Paper's Contribution Evidence \#1: \texttt{<evidence towards argument>} \\
Author 1's relevant evidence to potentially counter the quality of this contribution: \texttt{<counter evidence>} \\
... (more contributions and counter-evidence)
\\\\
Based on the debate history and your/your opposition's arguments and evidence, you must construct a new, stronger argument related to the topic. This consists of an argument that addresses/is robust to any doubts or clarifying questions made by the opposition which you feel are valid. If based on the debate, you feel that you do not contribute to the given topic or your contributions ARE NOT better than the opposition's, then state so by conceding to the opposition (e.g., 'I do not believe my paper makes a better contribution than yours') and explain why.
\\\\
conversation\_history: \texttt{<conversation history>}
 \\ \hline
\end{tabularx}
\caption{Persona prompt to \textit{revise} to an argument.}
\label{prompt_persona: revise_argument}
\end{table*}

% \begin{table*}[h]
% \centering
% \begin{tabularx}{\textwidth}{|X|}
% \hline
%  \\ \hline
% \end{tabularx}
% \caption{}
% \label{prompt}
% \end{table*}

\end{document}